\documentclass[10pt,twocolumn,letterpaper]{article}

\usepackage{cvpr}              
\definecolor{cvprblue}{rgb}{0.21,0.49,0.74}
\usepackage[accsupp]{axessibility}  
\usepackage[pagebackref,breaklinks,colorlinks,allcolors=cvprblue]{hyperref}

\usepackage{amsmath,amsfonts,bm}

\usepackage{mdframed}
\definecolor{theoremcolor}{rgb}{0.94, 0.94, 0.94}
\definecolor{examplecolor}{rgb}{1, 1, 1.0}
\mdfsetup{
    backgroundcolor=theoremcolor,
    linewidth=0pt,
}
\usepackage{xfrac}
\usepackage{comment}
\newmdtheoremenv[linewidth=0pt,innerleftmargin=4pt,innerrightmargin=4pt]{prop}{Proposition}
\newmdtheoremenv[linewidth=0pt,innerleftmargin=4pt,innerrightmargin=4pt]{assump}{Assumption}
\newmdtheoremenv[linewidth=0pt,innerleftmargin=4pt,innerrightmargin=4pt]{defn}{Definition}
\newmdtheoremenv[linewidth=0pt,innerleftmargin=4pt,innerrightmargin=4pt]{theorem}{Theorem}
\newmdtheoremenv[linewidth=0pt,innerleftmargin=4pt,innerrightmargin=4pt]{lemma}{Lemma}

\usepackage{amsmath}
\usepackage{amssymb}

\newcommand{\rvt}{{\mathbf{t}}}
\newcommand{\rvv}{{\mathbf{v}}}
\newcommand{\rvf}{{\mathbf{f}}}

\newcommand{\vframes}{{\mathbf{F}}}

\newcommand{\vtext}{{\rvt}}
\newcommand{\vtexti}[1]{{\rvt_{#1}}}

\newcommand{\vstotexti}[1]{{\rvt_{#1}^{\text{sto}}}}

\newcommand{\vvideo}{{\rvv}}

\newcommand{\eps}{\bm{\epsilon}}

\newcommand{\loss}{\mathcal{L}}

\usepackage{mathtools}
\DeclarePairedDelimiterX{\infodivx}[2]{(}{)}{%
	#1\;\delimsize\|\;#2%
}

\DeclareMathOperator*{\argmax}{{\arg\max}}

\usepackage{algorithmic}
\usepackage{xspace}
\usepackage{booktabs}       
\usepackage{amsfonts}       
\usepackage{nicefrac}       
\usepackage{microtype}      
\usepackage{xcolor}         
\usepackage{bm}
\usepackage{multirow}
\usepackage{threeparttable}
\usepackage{wrapfig}
\usepackage{graphicx}
\usepackage{makecell}
\usepackage{color}
\usepackage{algorithm}
\usepackage[capitalize]{cleveref}
\usepackage{colortbl}
\usepackage{lscape}
\usepackage{enumitem}
\usepackage[numbers]{natbib}

\newcommand{\appcref}[1]{Appendix~\ref{#1}}

\newcommand{\name}{Energy-Aware Fine-Grained Relationship Learning Network\xspace}
\newcommand{\bname}{\textbf{E}nergy-\textbf{A}ware Fine-\textbf{G}rained Relationship \textbf{Le}arning \textbf{Net}work\xspace}
\newcommand{\shortname}{EagleNet\xspace}

\newcommand{\g}{FRL\xspace}
\newcommand{\e}{EAM\xspace}
\newcommand{\esup}{eam\xspace}

\newcommand{\ct}[1]{\cite{#1}}

\def\paperID{****} 
\def\confName{CVPR}
\def\confYear{2026}

\title{\shortname: Energy-Aware Fine-Grained Relationship Learning Network for Text-Video Retrieval}

\author{
\textbf{Yuhan Chen}$^1$ \quad
\textbf{Pengwen Dai}$^{2,3}$\thanks{Corresponding authors: Pengwen Dai (daipw@mail.sysu.edu.cn) and Xiaochun Cao (caoxiaochun@mail.sysu.edu.cn).} \quad
\textbf{Chuan Wang}$^4$ \quad
\textbf{Dayan Wu}$^5$ \quad
\textbf{Xiaochun Cao}$^{2,3*}$\\
$^{1}$Sun Yat-sen University \quad
$^{2}$Shenzhen Campus of Sun Yat-sen University \\
$^{3}$Shenzhen Key Laboratory of Adversarial Artificial Intelligence \\
$^{4}$Beijing Jiaotong University \quad
$^{5}$Institute of Information Engineering, CAS\\
Guangzhou, China$^1$ \quad Shenzhen, China$^{2,3}$ \quad Beijing, China$^{4,5}$\\
{\tt\small draym28@gmail.com \quad daipw@mail.sysu.edu.cn \quad wangchuan@bjtu.edu.cn}\\
{\tt\small wudayan@iie.ac.cn \quad caoxiaochun@mail.sysu.edu.cn}
\vspace{-2.8mm}
}

\begin{document}
\maketitle


\begin{abstract}

\vspace{-5mm}
Text-video retrieval tasks have seen significant improvements due to the recent development of large-scale vision-language pre-trained models. 
Traditional methods primarily focus on video representations or cross-modal alignment, while recent works shift toward enriching text expressiveness to better match the rich semantics in videos. 
However, these methods use only interactions between text and frames/video, and ignore rich interactions among the internal frames within a video, so the final expanded text cannot capture frame contextual information, leading to disparities between text and video. 
In response, we introduce \name (\shortname) to generate accurate and context-aware enriched text embeddings. 
Specifically, the proposed Fine-Grained Relationship Learning mechanism (\g) first constructs a text-frame graph by the generated text candidates and frames, then learns relationships among texts and frames, which are finally used to aggregate text candidates into an enriched text embedding that incorporates frame contextual information. 
To further improve fine-grained relationship learning in \g, we design Energy-Aware Matching (\e) to model the energy of text-frame interactions and thus accurately capture the distribution of real text-video pairs. 
Moreover, for more effective cross-modal alignment and stable training, we replace the conventional softmax-based contrastive loss with the sigmoid loss.
Extensive experiments have demonstrated the superiority of \shortname across MSRVTT, DiDeMo, MSVD, and VATEX. 
Codes are available at \url{https://github.com/draym28/EagleNet}.

\end{abstract}
\vspace{-5mm}
\section{Introduction}

\begin{figure}
    \centering
    \includegraphics[width=0.8\linewidth]{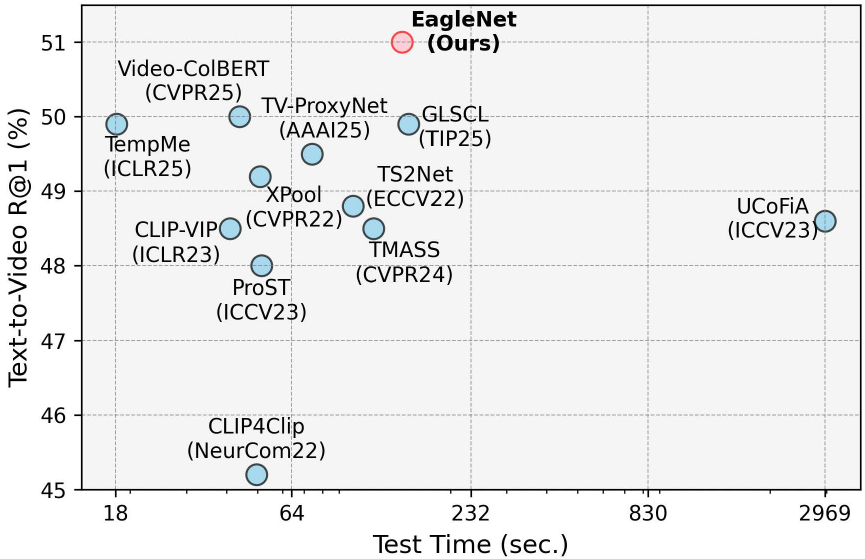}
    \vspace{-8pt}
    \caption{The performance-efficiency comparison among SOTA models (ViT-B/16) on MSRVTT datasets. 
    }
    \label{fig:r1-time}
    \vspace{-15pt}
\end{figure}

As the volume of video data being generated and stored continues to grow, the need for effective content retrieval models has become more urgent than ever.
Text-Video Retrieval (TVR) has been extensively studied and shows great progress in recent years. 
In Retrieval-Augmented Generation (RAG)~\cite{gao2023retrieval}, TVR improves the accuracy of generative models by integrating video content into knowledge-based systems, enhancing user interactions in applications like dialogue systems.

Early works~\ct{yu2018joint,croitoru2021teachtext,wray2019fine,dong2021dual,zhao2022centerclip,ge2022bridging,gabeur2020multi,liu2021hit,lei2021less,wang2021t2vlad,chen2020fine,liu2019use,fang2022concept} mainly rely on pre-extracted features, which are learned by independent text/image encoders. 
Additionally, due to the huge development in Graph Neural Network (GNN), many researchers utilize graph learning techniques in text-video retrieval tasks. 
As the groundbreaker, HGR~\ct{chen2020fine} constructs a semantic graph for each sentence, and utilizes GNN to learn text embeddings. 
Inspired by HGR, many follow-up works~\ct{wang2020learning,jin2021hierarchical,hao2021multi,fang2022concept} employ GNN for text-video structural relationship learning. 
For example, ACP~\ct{fang2022concept} introduces knowledge graph learning model to help complete semantic information for retrieval. 
However, all of these works are based on data pre-processing approaches with heavy external knowledge or outdated encoders, limiting the flexibility and increasing the complexity of the model. 

Instead, recent studies incorporate the emergent large-scale image-text pre-trained models into text-video retrieval tasks~\ct{luo2022clip4clip,gorti2022x,ma2022x,xue2023clip,wang2023unified,li2023progressive,wang2024text,fang2021clip2video,liu2022ts2,cheng2021improving,reddy2025video,xiao2025text,yu2025she,zhang2025text,cao2024rap,shentempme}. 
They utilize CLIP image and text encoder~\ct{radford2021learning} as multi-modal encoders to achieve strong performance on many text-video retrieval benchmarks. 
CLIP4Clip~\ct{luo2022clip4clip}, the first work that applies CLIP in this task, proposes using a temporal aggregator to fuse frames and then align text and video features. 
Later, many works~\ct{gorti2022x,fang2021clip2video,liu2022ts2,cheng2021improving,zhang2025text,cao2024rap,shentempme} focus on obtaining high-quality video features to enhance cross-modal matching.
Besides, some other works~\ct{zou2022tokenflow,li2023progressive,wang2023unified,yu2025she,zhang2025text,reddy2025video} propose to improve the text-video alignment, such as conducting matching at different granularities.
However, most models overlook the insufficient expressiveness of text embeddings. 
The brevity of video descriptions often prevents text embeddings from fully reflecting the underlying video semantics. 
To fill this gap, 
TMASS~\ct{wang2024text} stochastically samples enriched text within a flexible range, the radius of which is determined by the similarity of text and frame embeddings. 
Similarly, TV-ProxyNet~\ct{xiao2025text} refines the text query into a specific text proxy for each candidate video proxy through video-aware directors and text-video similarities. 
Nevertheless, these methods focus solely on interactions between text and frames/video when expanding text semantics, and ignore relationships among video frames, so the final enriched text cannot capture frame contextual information, leading to disparities between text and video. 

We argue that the internal frame-to-frame relations within the video should also be incorporated when expanding text semantics. 
This enables the model to capture both global and temporal semantics of videos, allowing the generated text embeddings to reflect not only the semantics of individual frames but also their contextual dependencies. 
Moreover, by establishing relationships among video semantic units (frames), the model can effectively identify and suppress redundant information and noise, leading to more accurate, contextually enriched text embeddings.

To this end, we introduce the \bname (\textbf{\shortname}), 
a framework designed to generate accurate and context-aware enriched text embeddings by exploring fine-grained relationships among text and video frames (see~\cref{fig:framework}). 
The model consists of two key components: 1) Fine-Grained Relationship Learning (\g) and 2) Energy-Aware Matching (\e). 
Specifically, \g first generates multiple text candidates using a stochastic text modeling strategy, and constructs a text-frame relational graph. 
We then design an effective relational graph attention network to learn the inter-connections among texts and frames. These connections incorporate both text-frame interactions and frame contextual information,  which are subsequently used to aggregate the text candidates into the final enriched text embedding. 
To further enhance fine-grained text-frame relationship learning in \g, \e employs energy-based model to accurately model the energy of detailed text-frame interactions and thus capture the true distribution of text-video pairs, improving the overall matching performance. 
Moreover, we proposed replacing the conventional softmax-based contrastive loss~\ct{oord2018representation} with sigmoid loss~\ct{zhai2023sigmoid} for more effective cross-modal alignment and stable training. 
Softmax loss normalizes the logits across both dimensions of the batch similarity matrix, making it sensitive to the negative samples and batch size \ct{chuang2020debiased,karpukhin2020dense,zhai2023sigmoid}. 
Instead, sigmoid loss treats each pair independently, so it can mitigate these drawbacks and is naturally suitable for the multi-matched scenario of TVR, where one text may be semantically similar to multiple videos or vice versa. 
Extensive experiments have been conducted on various benchmarks, 
demonstrating our effectiveness and superiority.

Our main contributions can be summarized as follows: 
\begin{itemize}
    \item We design a fine-grained relationship learning network that considers not only text-video interaction but also internal frame-to-frame relations within videos, generating accurate enriched text embedding that incorporates frames' contextual information; 
    \item To the best of our knowledge, we are the first to integrate EBM into TVR, which effectively models fine-grained text-frame interactions, enhancing the relational learning and cross-modal alignment. 
    \item Extensive experiments illustrate the superiority of \shortname and the effectiveness of each component, showing improvement over state-of-the-art models across four benchmark datasets.
\end{itemize}
\section{Related Work}

\noindent\textit{\textbf{Text-Video Retrieval.}}
Text-Video Retrieval (TVR) needs to align text embeddings with video embeddings~\ct{hanu2022vtc,wray2021semantic,wang2022multi,hu2022lightweight}. 
Early works are mostly offline models that focus on designing a matching algorithm for pre-extracted features~\ct{yu2018joint,croitoru2021teachtext,wray2019fine,dong2021dual,zhao2022centerclip,ge2022bridging,gabeur2020multi,liu2021hit,lei2021less,wang2021t2vlad,chen2020fine,liu2019use,fang2022concept}. 
JPoSE~\ct{wray2019fine} and DualEncoding~\ct{dong2021dual} extract video-text features offline, then perform embedding alignment in a common space. 
To enhance representation quality, CE~\ct{liu2019use}, MMT~\ct{gabeur2020multi}, and HiT~\ct{liu2021hit} employ multiple unimodal experts (OCR, speech, etc.) to enrich video features. 
TeachText~\ct{croitoru2021teachtext} adopts multiple text encoders to gather text information and obtain more credible alignments. 
Besides, HGR~\ct{chen2020fine} and T2vlad~\ct{wang2021t2vlad} build semantic graphs or shared centers to explore global-local relations. 
However, these models employ additional analysis tools or experts to pre-process data, which increases the complexity and limits flexibility. 

Recently, more and more researchers aim to seek end-to-end solutions~\ct{luo2022clip4clip,gorti2022x,ma2022x,xue2023clip,wang2023unified,li2023progressive,wang2024text,fang2021clip2video,liu2022ts2,cheng2021improving,lei2021less,ge2022bridging,zhao2022centerclip}. 
ClipBERT~\ct{lei2021less} and Frozen~\ct{bain2021frozen} first propose efficient end-to-end pre-trained schemes. 
MCQ~\ct{ge2022bridging} builds cross-modal associations by predicting verb or noun features. 
Later, benefiting from the large-scale vision-language pre-trained models~\ct{radford2021learning}, 
CLIP4clip~\ct{luo2022clip4clip} uses CLIP as the backbone network, significantly improving TVR performance, and inspiring a series of works~\ct{liu2022ts2,gorti2022x,zhao2022centerclip,fang2021clip2video,jin2022expectation,cao2024rap,shen2025temporal}. 
Centerclip~\ct{zhao2022centerclip} conducts segment-level clustering, which reduces token redundancy and computational overhead. 
TS2Net~\ct{liu2022ts2} designs the token shift module to realize the perception of local movement between frames. 
CLIP-VIP~\ct{xue2023clip} learns video proxies to incorporate frame context. 
XPool~\ct{gorti2022x} uses text as a condition to guide the aggregation of video tokens. 
UCoFiA~\ct{wang2023unified} conducts hierarchical alignment to achieve coarse- and fine-grained matching. 
Similarly, GLSCL~\ct{zhang2025text} proposes a global and local interaction module for coarse- and fine-grained semantic consistent learning. 
ProST~\ct{li2023progressive} extracts higher-level prototypes for texts and videos and conducts spatio-temporal matching.
TempMe~\ct{shentempme} introduces a multi-granularity framework to progressively combine neighboring clips, reducing video redundancy and enhancing temporal modeling across frames. 
Video-ColBERT~\ct{reddy2025video} utilizes spatiotemporal token interactions, learnable query and visual expansion, and sigmoid loss training.

However, most existing works focus on learning high-quality video embeddings or improving multi-modal alignment, while overlooking the limited expressiveness of textual representations. 
TMASS~\ct{wang2024text} extends text embeddings effectively by sampling stochastic text embeddings within a flexible range, whose radius is determined by the similarity between texts and videos. 
Similarly, TV-ProxyNet~\ct{xiao2025text} converts the text query to a specific text proxy for each candidate video proxy through video-aware directors and text-video similarities.
Nevertheless, these methods only consider text-video interactions while expanding text semantics, ignoring the internal frame-to-frame relations within a video, so the final enriched text cannot capture frame context, leading to disparities between text and video.

\textit{\textbf{Graph Neural Networks.}}
Graph Neural Network (GNN) has received significant interest in recent years due to its powerful ability in real-world applications based on graph-structured data~\ct{kipf2017semi,velivckovic2018graph,schlichtkrull2018modeling}. 
The Graph Attention Network (GAT)~\ct{velivckovic2018graph} is the first to introduce the attention mechanism into GNN, and it shows great potential in modeling relationships. 
Furthermore, to tackle complex real-life problems related to heterogeneous graphs, such as processing knowledge graphs, Relational GNN (RGNN)~\ct{schlichtkrull2018modeling} and many successors have been proposed.
HGR~\ct{chen2020fine} is the first work to employ GNN in TVR; it constructs semantic graphs for texts by grammar and extracts representations using GNN. 
Later, many following works are proposed to employ GNN in TVR~\ct{wang2020learning,jin2021hierarchical,hao2021multi}. 
ACP~\ct{fang2022concept} uses knowledge graphs as prior knowledge to complete the semantic information. 
SHE-Net~\ct{yu2025she} introduces CLIP as the backbone and uses hierarchical text graphs to guide the aggregation of temporal frames and spatial regions. 
But these works are based on pre-extracted features and heavily rely on external knowledge, limiting the flexibility and increasing model complexity.

Unlike previous methods, our approach captures both the text-video interactions and rich relationships among video frames, which are essential for comprehending the video context. 
By leveraging these fine-grained interactions, the model is able to generate high-quality, context-aware text embeddings and achieve more accurate text-video alignment.

\section{Method}

\begin{figure*}
    \centering
    \includegraphics[width=0.85\linewidth]{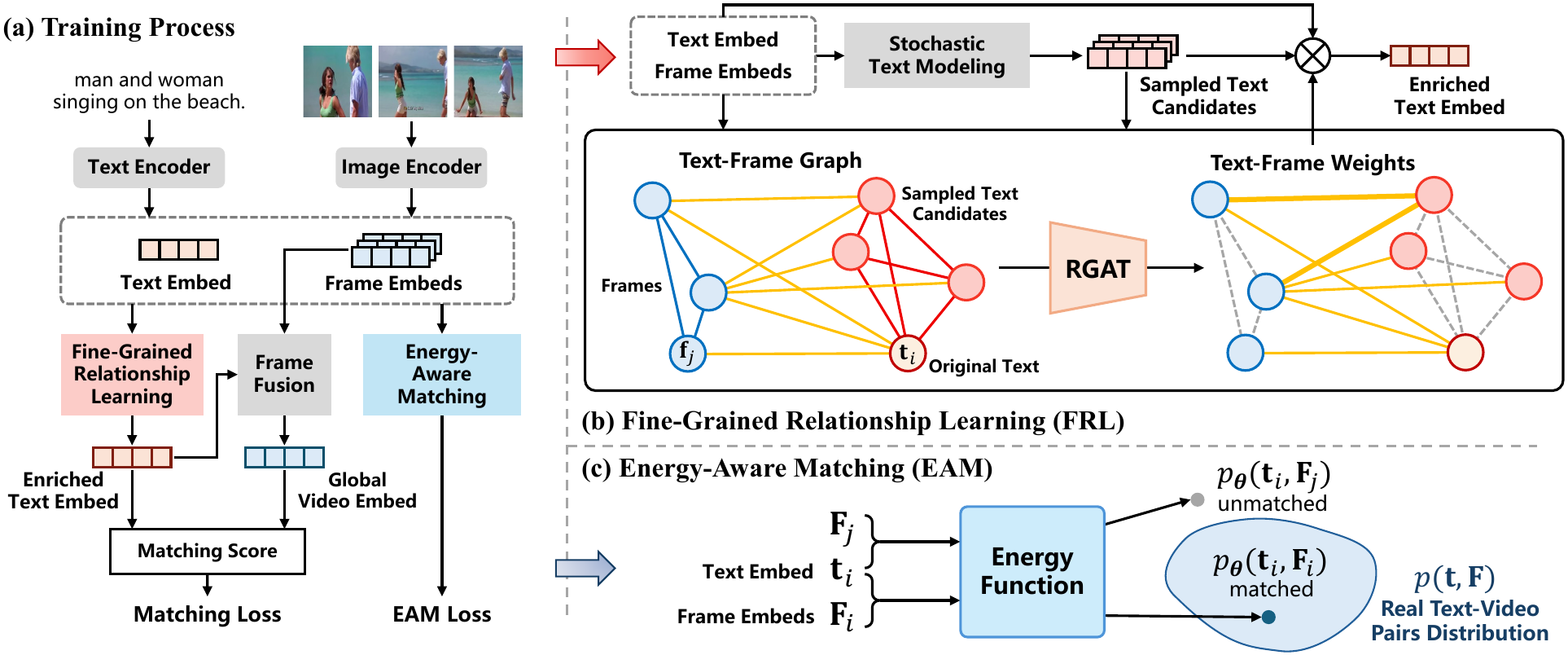}
    \vspace{-8pt}
    \caption{
    Diagram of \shortname. 
    (a) Overview of the training process. 
    (b) Fine-Grained Relationship Learning (\g) first samples multiple text candidates, then constructs a text-frame graph to learn both text-frame and frame-frame relationships, which are then used to aggregate text candidates into a final enriched text embedding aware of video context information.  
    (c) Energy-Aware Matching (\e) improves the relationships learning in \g by capturing detailed text-frame interactions and accurately models true text-video pairs distribution, thereby also enhancing the final matching performance from a fine-grained perspective. 
    }
    \label{fig:framework}
    \vspace{-10pt}
\end{figure*}

\subsection{Preliminary}

We first introduce the standard pipeline of the text-video retrieval task, 
including: 1) obtaining representations of text and video; 2) training and inference. 

\subsubsection{Text and Video Representations}

Text-video retrieval task aims to align the text and video in a joint space. 
The first step is to learn the text and video embeddings using the vision-language pre-trained model CLIP~\ct{radford2021learning}. 
Specifically, given a text-video pair $\{t,v\}$, TVR first samples $M$ frames from $v$ uniformly, $[f_1,f_2,\cdots,f_M]$, and then feeds them into CLIP text encoder $\phi_{\text{t}}(\cdot)$ and image encoder $\phi_{\text{v}}(\cdot)$, 
obtaining text embedding $\rvt$ and frame embeddings $\vframes$ as:
\begin{small}
\begin{equation}
    \begin{aligned}
        \rvt &= \phi_{\text{t}}(t) \in \mathbb{R}^{d}, \\
        \vframes &= \phi_{\text{v}}([f_1,f_2,\cdots,f_M])=[\rvf_1\|\rvf_2\|\cdots\|\rvf_M]^{\top} \in \mathbb{R}^{M \times d}, 
    \end{aligned}
\end{equation}
\end{small}
where ``$\|$" is the column concatenation operation, and $d$ is the hidden dimension of embeddings. 

Then a frame fusion module is employed to obtain the global video feature $\rvv=\pi(\vframes) \in\mathbb{R}^d$, to reduce the temporal redundancy in video and align $\rvt$ with $\rvv$. 
A simple solution is average pooling. But it treats all the frames equally, overlooking the important frames and retaining noise information from irrelevant frames. 
We follow XPool~\ct{gorti2022x}~(\appcref{sec:app.video_pooling}) to use $\rvt$ as condition, such that the fusion module can attach more attention to the text-related frames:
\begin{small}
\begin{equation}
    \rvv = \pi( \vframes | \rvt ) \in \mathbb{R}^d,
\label{eq:fuse_f_with_t}
\end{equation}
\end{small}
which uses cross-attention to fuse frames guided by text.

\subsubsection{Training and Inference}

In training stage, for each training batch $\{(t_i, v_i)\}_{i=1}^{B}$ with batch size $B$, 
a symmetric cross-entropy loss $\loss_{\text{ce}}$ is applied to minimize the distance of relevant text-video pairs while maximizing that between irrelevant pairs, similar to contrastive learning \ct{oord2018representation}. 
We consider both text-to-video retrieval ($\loss_{\text{ce}}^{t2v}$) and video-to-text retrieval ($\loss_{\text{ce}}^{v2t}$):
\begin{small}
\begin{equation}
    \begin{aligned}
        \loss_{\text{ce}}(\rvt, \rvv) &= \frac{1}{2}\Big(\loss_{\text{ce}}^{t2v}(\rvt, \rvv)+\loss_{\text{ce}}^{v2t}(\rvt, \rvv)\Big), \\
        \loss_{\text{ce}}^{t2v}(\rvt, \rvv) &= 
        -\frac{1}{B} \sum_{i=1}^{B} \log \frac{
        e^{ \operatorname{s}(\rvt_{i}, \rvv_{i})\cdot\lambda }}{
        \sum_{j=1}^{B} e^{ \operatorname{s}(\rvt_{i}, \rvv_{j})\cdot\lambda }}, \\
        \loss_{\text{ce}}^{v2t}(\rvt, \rvv) &= 
        -\frac{1}{B} \sum_{i=1}^{B} \log \frac{
        e^{ \operatorname{s}(\rvt_{i}, \rvv_{i})\cdot\lambda }}{
        \sum_{j=1}^{B} e^{ \operatorname{s}(\vtexti{j}, \rvv_{i})\cdot\lambda }},
    \end{aligned}
\label{eq:ce_loss}
\end{equation}
\end{small}
where $\operatorname{s}(\cdot,\cdot)$ is similarity function (\eg, cosine similarity) and $\lambda$ is a learnable factor. 

In inference, given $n_t$ texts $\{t_i\}_{i=1}^{n_t}$ and $n_v$ video clips $\{v_j\}_{j=1}^{n_v}$, 
for a query text $t_{\text{query}}$, the matched video clip $v^*$ is the one that has the highest matching score and vice versa:
\begin{small}
\begin{equation*}
    v^* = \argmax_{j} \operatorname{s}(\rvt_{\text{query}}, \rvv_{j}), \ \ \ t^* = \argmax_{i} \operatorname{s}(\rvt_{i}, \rvv_{\text{query}}).
\end{equation*}
\end{small}

\subsection{Our Proposed Model: \shortname}

In this section we introduce \textbf{\shortname} (\cref{fig:framework}), which comprises two key components: 
1) Fine-Grained Relationship Learning (\textbf{\g}) and 2) Energy-Aware Matching (\textbf{\e}).
First, \g generates context-aware enriched text by sampling text candidates and extracting fine-grained relationships among texts and frames (\cref{sec:frl}). 
Then, \e enhances \g in learning text-frame interactions and improves the final matching from a fine-grained view (\cref{sec:eam}). 
Finally, the sigmoid function is employed in the final loss for more effective alignment and stable training (\cref{sec:sigloss}).

\subsubsection{Fine-Grained Relationship Learning}\label{sec:frl}

Existing models only consider text-frame interactions when expanding text semantics. 
However, the internal frame-frame relations are also important to capture both global and temporal semantics. 
Specifically, we first sample multiple text candidates, 
and the texts and frames are treated as nodes to form a text-frame graph, which includes both text-frame and frame-frame relations. 
Then the relations are extracted by relational graph learning and used to fuse text candidates, generating text embedding that reflects not only the semantics of individual frames but also their contextual dependencies. 

\textit{\textbf{Text-Frame Graph Construction.}}
Given text embedding $\vtext$ and frame embeddings $\vframes=[ \rvf_1\|\rvf_2\|\cdots\|\rvf_M ]^{\top}$, 
we first sample $S$ stochastic text candidates $\{ \vstotexti{i} \}_{i=1}^S$ using stochastic text modeling~\ct{wang2024text} (\appcref{sec:app.sto_txt_modeling}), 
then concatenate texts and frames to get the node matrix as:
\begin{small}
\begin{equation}
\begin{aligned}
    \mathbf{X} &= \left[ \vtext\|\vstotexti{1}\|\vstotexti{2}\|\cdots\|\vstotexti{S}\|\rvf_1'\|\rvf_2'\|\cdots\|\rvf_M' \right]^{\top} \in \mathbb{R}^{n \times d}, \\
    \rvf_j' &= \rvf_j + \operatorname{PE}(j), 
\end{aligned}
\end{equation}
\end{small}
where $n=1+S+M$ is the number of nodes, and $\operatorname{PE}(\cdot)$ is learnable temporal positional embedding on frames. 
Thus, there are three types of edges (relations): text-text, frame-frame, and text-frame, denoted by $r \in \{ tt, ff, tf \}$. 

\textit{\textbf{Relationship Learning.}}
To extract useful information from the complex relations among texts and frames, 
we propose a simple yet effective version of the Relational Graph Attention Network (\textbf{RGAT}) with $L$ layers. 
The layer index is omitted here for brevity. 
In each layer, for any node pair $(i,j)$ linked by relation $r$, the edge weight is given by
\begin{small}
\begin{align}
    e_{ij}^{r,h} &= \psi^r \left( \left[\mathbf{W}^{r,h}\mathbf{h}_i \| \mathbf{W}^{r,h}\mathbf{h}_j \right] \right) \in \mathbb{R}, 
\label{eq:eij_r}
\end{align}
\end{small}
where $h=1,2,\cdots,H$ is the indices of attention head, $\mathbf{h}_i$ is the feature of node $i$ (in the first layer $\mathbf{h}_i$ is the $i$-th row of $\mathbf{X}$), 
$\mathbf{W}^{r,h}$ is learnable matrix, and $\psi^r(\cdot): \mathbb{R}^{2d} \mapsto \mathbb{R}$ is projector to obtain edge weights. 
Then, the attention score is obtained as
\begin{small}
\begin{align}
    \alpha_{ij}^{r,h} &= \frac{ \exp \left(\operatorname{LeakyReLU}(e_{ij}^{r,h} )\right) }{ \sum_{k\in\mathcal{N}_i^r}\exp\left(\operatorname{LeakyReLU}(e_{ik}^{r,h} )\right) }, 
\label{eq:attns_r}
\end{align}
\end{small}
where $\mathcal{N}_i^r$ is the neighbor set of node $i$ linked by relation $r$.
Finally, the output node representation of this layer $\mathbf{h}_i'$ is
\begin{small}
\begin{equation}
    \mathbf{h}_i' = \operatorname{ReLU} \Big( \mathbf{W}_{\text{out}} \mathbf{h}_i +
    \Big\|_{h=1}^{H} \sum_r\sum_{j\in\mathcal{N}_i^{r}}\alpha_{ij}^{r,h}\mathbf{W}^{r,h}\mathbf{h}_j \Big), 
\label{eq:rgat}
\end{equation}
\end{small}
where $\mathbf{W}_{\text{out}}\in\mathbb{R}^{Hd\times d}$ is learnable weights for residual connection. Then $\mathbf{h}_i'$ is input of the next layer. 
Matrix form and more details can be found in Appendix~\ref{sec:app.rgat}. 

Note that here $\mathbf{h}_i'$ only serves as an intermediate. 
The target of \g is text-frame relations $e_{ij}^{tf}$, which are regarded as the importance of text embeddings for the frames. 
These weights are then used to aggregate text embeddings into a final enriched text. 
Specifically, the text-frame edge weights from \textbf{the final layer} are averaged across attention heads to get the final text-frame weights, \ie, 
\begin{small}
\begin{equation}
    e_{ij}^{tf} = \frac{1}{H} \sum_{h=1}^{H}e_{ij}^{tf,h}, \ \ 
    \begin{aligned}
        &i \in [1, 1+S], \\
        &j \in [1+S+1, 1+S+M], 
    \end{aligned}
\label{eq:text_frame_weights}
\end{equation}
\end{small}
then we average and normalize the text-frame weights to obtain the attention score of each text node:
\begin{small}
\begin{equation}
    \begin{aligned}
        w_i = \frac{ \exp ( e_i^{tf} ) }{ \sum_{k=1}^{1+S} \exp ( e_k^{tf} ) }, \ \ \  
        e_i^{tf} = \frac{1}{M} \sum_{j=1+S+1}^{1+S+M} e_{ij}^{tf}, 
    \end{aligned}
\label{eq:text_nodes_attns}
\end{equation}
\end{small}
so the final enriched text embedding is
\begin{small}
\begin{equation}
    \rvt^{\text{gen}} = \sum_{i=1}^{1+S} w_i \mathbf{X}_i = w_1\vtext +\sum_{i=1}^{S} w_{i+1}\vstotexti{i} \in \mathbb{R}^{d}. 
\label{eq:text_gen}
\end{equation}
\end{small}

$\rvt^{\text{gen}}$ is used as the condition for frame fusion (\cref{eq:fuse_f_with_t}) to obtain video embedding $\rvv$, 
then $\loss_{\text{ce}}(\rvt^{\text{gen}}, \rvv)$ is calculated (\cref{eq:ce_loss}). 
Following~\ct{wang2024text}~(\appcref{sec:app.sto_txt_modeling}), we further obtain a support text $\rvt^{\text{sup}}$, and calculate the support loss $\loss_{\text{ce}}(\rvt^{\text{sup}}, \rvv)$. 
$\rvt^{\text{sup}}$ serves as the boundary of the stochastic modeling range, within which $\vstotexti{i}$ is sampled. 
So the matching loss is
\begin{small}
\begin{equation}
    \loss_{\text{match}} = \loss_{\text{ce}}(\rvt^{\text{gen}}, \rvv) + \lambda_{\text{sup}} \loss_{\text{ce}}(\rvt^{\text{sup}}, \rvv), 
\label{eq:match_loss}
\end{equation}
\end{small}
where $\lambda_{\text{sup}}$ is a coefficient.

\subsubsection{Energy-Aware Matching}\label{sec:eam}

\begin{table*}[!t]
\centering

\resizebox{0.8\linewidth}{!}{

\begin{tabular}{cl|ccccc|ccccc|c}
\multicolumn{2}{c|}{\multirow{2}{*}{\textbf{Method}}} & \multicolumn{5}{c|}{\textbf{Text-to-Video}} & \multicolumn{5}{c|}{\textbf{Video-to-Text}} & \multirow{2}{*}{\textbf{Rsum↑}} \\
\multicolumn{2}{c|}{} & \textbf{R@1↑} & \textbf{R@5↑} & \textbf{R@10↑} & \textbf{MdR↓} & \textbf{MnR↓} & \textbf{R@1↑} & \textbf{R@5↑} & \textbf{R@10↑} & \textbf{MdR↓} & \textbf{MnR↓} &  \\
\hline
\multicolumn{13}{c}{\textbf{Not End-to-End Method}} \\
\hline
\multirow{5}{*}{} & CE~\ct{liu2019use} \textcolor{gray!60}{BMVC19} & 20.6 & 50.3 & 64.0 & 5.3 & - & 20.9 & 48.8 & 62.4 & 6.0 & - & 267.0 \\
 & MMT~\ct{gabeur2020multi} \textcolor{gray!60}{ECCV20} & 26.6 & 57.1 & 69.6 & 4.0 & 24.0 & 27.0 & 57.5 & 69.7 & 3.7 & 21.3 & 307.5 \\
 & T2vlad~\ct{wang2021t2vlad} \textcolor{gray!60}{CVPR21} & 29.5 & 59.0 & 70.1 & 4.0 & - & 31.8 & 60.0 & 71.1 & 3.0 & - & 321.5 \\
 & HiT~\ct{liu2021hit} \textcolor{gray!60}{ICCV21} & 27.7 & 59.2 & 72.0 & 2.9 & - & 28.8 & 60.3 & 72.3 & 3.0 & - & 320.3 \\
 & TeachText~\ct{croitoru2021teachtext} \textcolor{gray!60}{ICCV21} & 29.6 & 61.6 & 74.2 & 3.0 & - & 32.1 & 62.7 & 75.0 & 3.0 & - & 335.2 \\
\hline
\multicolumn{13}{c}{\textbf{End-to-End Method}} \\
\hline
\multirow{5}{*}{\rotatebox{90}{\textit{non-CLIP}}} & ClipBERT~\ct{lei2021less} \textcolor{gray!60}{CVPR21} & 22.0 & 46.8 & 59.9 & 6.0 & - & - & - & - & - & - & - \\
 & SupportSet~\ct{patricksupport} \textcolor{gray!60}{ICLR21} & 30.1 & 58.5 & 69.3 & 3.0 & - & 28.5 & 58.6 & 71.6 & 3.0 & - & 316.6 \\
 & Frozen~\ct{bain2021frozen} \textcolor{gray!60}{ICCV21} & 31.0 & 59.5 & 70.5 & 3.0 & - & - & - & - & - & - & - \\
 & BridgeFormer~\ct{ge2022bridging} \textcolor{gray!60}{CVPR22} & 37.6 & 64.8 & 75.1 & 3.0 & - & - & - & - & - & - & - \\
 & TVMM~\ct{lin2022text} \textcolor{gray!60}{NeurIPS22} & 36.2 & 64.2 & 75.7 & 3.0 & - & 34.8 & 63.8 & 73.7 & 3.0 & - & 348.4 \\
\hline
\multirow{14}{*}{\rotatebox{90}{\textit{ViT-B/32}}} & CLIP4Clip$^{*}$~\ct{luo2022clip4clip} \textcolor{gray!60}{NeurCom22} & 43.2 & 71.4 & 80.6 & \textbf{2.0} & 16.5 & 42.5 & 70.9 & 81.0 & \textbf{2.0} & 12.2 & 389.6 \\
 & XPool$^{\dagger*}$~\ct{gorti2022x} \textcolor{gray!60}{CVPR22} & 45.8 & 73.8 & 82.0 & \textbf{2.0} & 14.0 & \textbf{45.7} & 73.1 & 82.3 & \textbf{2.0} & 9.4 & 402.7 \\
 & TS2Net$^{*}$~\ct{liu2022ts2} \textcolor{gray!60}{ECCV22} & 45.4 & 72.9 & 82.9 & \textbf{2.0} & 14.2 & 44.8 & 72.6 & 81.1 & \textbf{2.0} & 10.3 & 399.7 \\
 & CLIP-VIP$^{\dagger*}$~\ct{xue2023clip} \textcolor{gray!60}{ICLR23} & 46.0 & 72.6 & 82.2 & \textbf{2.0} & 14.4 & 43.6 & 71.5 & 82.3 & \textbf{2.0} & 10.8 & 398.2 \\
 & ProST$^{*}$~\ct{li2023progressive} \textcolor{gray!60}{ICCV23} & 46.3 & 73.6 & 82.4 & \textbf{2.0} & \underline{12.9} & 45.0 & 72.9 & 82.2 & \textbf{2.0} & \textbf{8.9} & 402.4 \\
 & TMASS$^{\dagger*}$~\ct{wang2024text} \textcolor{gray!60}{CVPR24} & 43.4 & 73.2 & 82.1 & \textbf{2.0} & 14.3 & 45.0 & 73.2 & 82.7 & \textbf{2.0} & 9.5 & 399.6 \\
 & RAP~\ct{cao2024rap} \textcolor{gray!60}{ACL24} & 44.8 & 71.4 & 81.5 & - & 14.4 & 44.0 & 71.9 & 82.4 & - & 10.1 & 396.0 \\
 & TFVL~\ct{shen2025temporal} \textcolor{gray!60}{TNNLS25} & 45.2 & 70.4 & 80.1 & - & 15.2 & 45.5 & 71.5 & 81.0 & - & 11.0 & 393.7 \\
 & SHE-Net~\ct{yu2025she} \textcolor{gray!60}{TCSVT25} & 45.3 & \textbf{74.9} & \textbf{84.2} & \textbf{2.0} & \textbf{12.5} & 44.8 & \textbf{74.8} & \textbf{83.7} & \textbf{2.0} & \underline{9.1} & \underline{407.7} \\
 & GLSCL$^{*}$~\ct{zhang2025text} \textcolor{gray!60}{TIP25} & \textbf{47.2} & 73.5 & 82.7 & \textbf{2.0} & 13.1 & \underline{45.6} & 73.5 & 82.7 & \textbf{2.0} & 9.8 & 405.2 \\
 & TempMe$^{*}$~\ct{shentempme} \textcolor{gray!60}{ICLR25} & 43.3 & 71.2 & 81.2 & \textbf{2.0} & 15.3 & 45.0 & 71.5 & 80.8 & \textbf{2.0} & 11.0 & 393.0 \\
 & TV-ProxyNet$^{\dagger*}$~\ct{xiao2025text} \textcolor{gray!60}{AAAI25} & \underline{46.7} & 73.0 & 81.8 & \textbf{2.0} & 14.5 & 45.4 & 71.1 & 82.4 & \textbf{2.0} & 11.0 & 400.4 \\
 & Video-ColBERT$^{\dagger*}$~\ct{reddy2025video} \textcolor{gray!60}{CVPR25} & 45.9 & 71.6 & 82.7 & \textbf{2.0} & 13.7 & 42.1 & 70.4 & 81.1 & \textbf{2.0} & 10.8 & 393.8 \\
\rowcolor{gray!15}
\cellcolor{white} & \textbf{\shortname (Ours)} & \textbf{47.2} & \underline{74.8} & \underline{84.0} & \textbf{2.0} & 14.4 & \textbf{45.7} & \underline{73.7} & \underline{83.4} & \textbf{2.0} & 9.4 & \textbf{408.8} \\
\hline
\multirow{14}{*}{\rotatebox{90}{\textit{ViT-B/16}}} & CLIP4Clip$^{*}$~\ct{luo2022clip4clip} \textcolor{gray!60}{NeurCom22} & 45.2 & 72.2 & 81.4 & 2.0 & 15.0 & 42.9 & 70.9 & 80.6 & \textbf{2.0} & 11.6 & 393.2 \\
 & XPool$^{\dagger*}$~\ct{gorti2022x} \textcolor{gray!60}{CVPR22} & 49.2 & 73.9 & 82.6 & 2.0 & 13.6 & 48.0 & 74.9 & 82.9 & \textbf{2.0} & 9.7 & 411.5 \\
 & TS2Net$^{*}$~\ct{liu2022ts2} \textcolor{gray!60}{ECCV22} & 48.8 & 75.0 & \underline{85.0} & 2.0 & 13.0 & 46.4 & 75.4 & 85.3 & \textbf{2.0} & 8.6 & 415.9 \\
 & CLIP-VIP$^{\dagger*}$~\ct{xue2023clip} \textcolor{gray!60}{ICLR23} & 48.5 & 75.5 & 83.9 & 2.0 & 12.5 & 47.6 & 74.8 & 85.0 & \textbf{2.0} & 9.6 & 415.3 \\
 & UCoFiA$^{*}$~\ct{wang2023unified} \textcolor{gray!60}{ICCV23} & 48.6 & 75.5 & 83.8 & 2.0 & \textbf{11.7} & \underline{49.0} & 75.7 & 84.2 & \textbf{2.0} & 10.0 & 416.8 \\
 & ProST$^{*}$~\ct{li2023progressive} \textcolor{gray!60}{ICCV23} & 48.0 & 74.7 & 84.0 & 2.0 & 12.7 & 48.3 & 75.2 & 84.5 & \textbf{2.0} & 8.7 & 414.7 \\
 & TMASS$^{\dagger*}$~\ct{wang2024text} \textcolor{gray!60}{CVPR24} & 48.5 & 74.1 & 83.4 & 2.0 & 14.4 & 46.7 & 75.9 & \underline{85.6} & \textbf{2.0} & 9.1 & 414.2 \\
 & RAP~\ct{cao2024rap} \textcolor{gray!60}{ACL24} & 46.5 & 73.9 & 82.0 & - & 12.1 & 45.3 & \underline{76.4} & 84.8 & - & 9.1 & 408.9 \\
 & TFVL~\ct{shen2025temporal} \textcolor{gray!60}{TNNLS25} & 48.7 & 73.5 & 83.0 & - & 13.4 & 47.3 & 75.3 & 83.7 & - & 9.5 & 411.5 \\
 & GLSCL$^{*}$~\ct{zhang2025text} \textcolor{gray!60}{TIP25} & 49.9 & \textbf{76.3} & 84.1 & 2.0 & \underline{11.9} & 48.3 & 75.8 & 84.6 & \textbf{2.0} & \textbf{8.1} & \underline{419.0} \\
 & TempMe$^{*}$~\ct{shentempme} \textcolor{gray!60}{ICLR25} & 49.9 & 75.3 & 84.8 & 2.0 & 12.1 & 47.5 & 75.2 & 85.1 & \textbf{2.0} & 8.8 & 417.8 \\
 & TV-ProxyNet$^{\dagger*}$~\ct{xiao2025text} \textcolor{gray!60}{AAAI25} & 49.5 & 74.9 & 83.4 & 2.0 & 12.6 & 47.9 & 74.8 & 84.1 & \textbf{2.0} & 9.4 & 414.6 \\
 & Video-ColBERT$^{\dagger*}$~\ct{reddy2025video} \textcolor{gray!60}{CVPR25} & \underline{50.0} & \textbf{76.3} & 84.3 & \underline{1.5} & 13.1 & 47.9 & 74.4 & 84.9 & \textbf{2.0} & 9.0 & 417.8 \\
\rowcolor{gray!15}
\cellcolor{white} & \textbf{\shortname (Ours)} & \textbf{51.0} & \underline{76.2} & \textbf{85.6} & \textbf{1.0} & 12.8 & \textbf{49.2} & \textbf{77.1} & \textbf{86.6} & \textbf{2.0} & \underline{8.2} & \textbf{425.7}
\end{tabular}

}

\vspace{-8pt}
\caption{Performance on MSRVTT, \textbf{boldface} is the best results and \underline{underline} is the runner-ups. 
Note that we have resolved the data-leakage problem in the original TMASS codes, 
and re-implemented XPool$^{\dagger}$, CLIP-VIP$^{\dagger}$, TMASS$^{\dagger}$, TV-ProxyNet$^{\dagger}$, and Video-ColBERT$^{\dagger}$ using CLIP4Clip codebase to align with other methods. 
$^{*}$ denotes the reproduced results, all of which are fine-tuned on ViT-B/32 or ViT-B/16. 
}
\label{tab:main_msrvtt9k}

\vspace{-10pt}
\end{table*}

Since $\loss_{\text{match}}$ only aligns text and video at global level, 
we propose \e to assist \g in learning fine-grained text-frame relations. 
\e employs energy-based model to effectively capture data distributions~\ct{du2019implicit,liu2020energy,chendecoupled},
thereby enhancing the matching from a fine-grained perspective. 

To be specific, given text embedding $\vtext$ and frame embeddings $\vframes=[\rvf_1 \| \cdots \| \rvf_M]^{\top}$, \e models the real text-video pair distribution $p(\vtext,\vframes)$ using Boltzmann distribution $p_{\theta}(\vtext,\vframes) = \frac{\exp{(-E_\theta(\vtext,\vframes))}}{Z_\theta}$, where $Z_\theta$ is the normalizing constant, $\theta$ is learnable parameter, and $E_{\theta}(\vtext,\vframes)\in\mathbb{R}$ denotes the text-video energy. 
We first calculate the energy for each text-frame pair and then average them to obtain the text-video energy $E_{\theta}(\vtext,\vframes)=\frac{1}{M}\sum_i^ME_{\theta}(\vtext,\rvf_i)$. 
Such that \e can fully utilize the information in fine-grained text-frame interactions.
Simultaneously, $E_{\theta}$ assigns low energy scores to the true text-video pairs and high energy to mismatched ones, 
further assisting the model in achieving more precise text-video alignment.

Additionally, besides averaging text-frame energies (\texttt{Avgpool}), other pooling approaches, \eg, \texttt{Maxpool}, are also options. This will be discussed in~\cref{sec:pooling_energy}. 

\textit{\textbf{Energy Functions.}}
Since the energy score indicates the relevance of a text-video pair, we can adopt a similarity function as $E_\theta$, 
such as negative cosine similarity (\texttt{CosSim}): 
\vspace{-2mm}
\begin{small}
\begin{equation}
    E(\vtext,\vframes)= \frac{1}{M} \sum_{i=1}^M E(\vtext,\rvf_i) = -\frac{1}{M} \sum_{i=1}^M \frac{\vtext^{\top}\rvf_i}{\|\vtext\|\|\rvf_i\|}. 
\label{eq:efunc_cossim}
\end{equation}
\end{small}
Or we can use the bilinear scoring function with learnable parameters $\mathbf{W}$ (\texttt{Bilinear}): 
\vspace{-2mm}
\begin{small}
\begin{equation}
    E_{\theta}(\vtext,\vframes) = -\frac{1}{M} \sum_{i=1}^M \frac{{\vtext}^{\top} \mathbf{W} \rvf_i}{\|\vtext\|\|\rvf_i\|}. 
\label{eq:efunc_bilinear}
\end{equation}
\end{small}
For greater flexibility, MLP is also a strong choice (\texttt{MLP}): 
\vspace{-2mm}
\begin{small}
\begin{equation}
    E_{\theta}(\vtext,\vframes) = \frac{1}{M} \sum_{i=1}^M \operatorname{MLP}([\vtext \| \rvf_i]). 
\label{eq:efunc_mlp}
\end{equation}
\end{small}

\begin{table*}[!t]
\centering

\resizebox{0.75\linewidth}{!}{

\begin{tabular}{cl|ccc|ccc|ccc|c}
\multicolumn{2}{c|}{\multirow{2}{*}{\textbf{Method}}} & \multicolumn{3}{c|}{\textbf{DiDeMo}} & \multicolumn{3}{c|}{\textbf{MSVD}} & \multicolumn{3}{c|}{\textbf{VATEX}} & \multirow{2}{*}{\textbf{Rsum↑}} \\
\multicolumn{2}{c|}{} & \textbf{R@1↑} & \textbf{R@5↑} & \textbf{R@10↑} & \textbf{R@1↑} & \textbf{R@5↑} & \textbf{R@10↑} & \textbf{R@1↑} & \textbf{R@5↑} & \textbf{R@10↑} &  \\
\hline
\multirow{7}{*}{} & CE~\ct{liu2019use} & 16.1 & 41.1 & 82.7 & 19.8 & 49.0 & - & - & - & - & - \\
 & HGR~\ct{chen2020fine} & - & - & - & - & - & - & 35.1 & 73.5 & 83.5 & - \\
 & ClipBERT~\ct{lei2021less} & 20.4 & 48.0 & 60.8 & - & - & - & - & - & - & - \\
 & SupportSet~\ct{patricksupport} & - & - & - & 28.4 & 60.0 & - & 45.9 & 82.4 & - & - \\
 & Frozen~\ct{bain2021frozen} & 31.0 & 59.8 & 72.4 & 33.7 & 64.7 & - & - & - & - & - \\
 & BridgeFormer~\ct{ge2022bridging} & 37.0 & 62.2 & 73.9 & - & - & - & - & - & - & - \\
 & TVMM~\ct{lin2022text} & 36.5 & 64.9 & 75.4 & - & - & - & - & - & - & - \\
\hline
\multirow{14}{*}{\rotatebox{90}{\textit{ViT-B/32}}} & CLIP4Clip$^{*}$~\ct{luo2022clip4clip} & 40.6 & 67.7 & 77.8 & 46.2 & 75.6 & 84.6 & 56.6 & 87.5 & 93.6 & 630.2 \\
 & XPool$^{\dagger*}$~\ct{gorti2022x} & 43.5 & 69.5 & 79.0 & 47.1 & 76.5 & 85.2 & 57.5 & 87.8 & 93.7 & 639.8 \\
 & TS2Net$^{*}$~\ct{liu2022ts2} & 39.5 & 69.7 & 80.9 & 44.7 & 75.5 & 84.9 & 57.7 & 88.7 & 94.2 & 635.8 \\
 & CLIP-VIP$^{\dagger*}$~\ct{xue2023clip} & 44.7 & 72.3 & 81.3 & 46.5 & 76.2 & 85.2 & \textbf{59.3} & 89.0 & \underline{94.3} & 648.8 \\
 & ProST$^{*}$~\ct{li2023progressive} & 45.8 & 74.3 & \textbf{84.5} & 44.3 & 75.2 & 84.8 & 57.3 & 88.1 & 94.2 & 648.5 \\
 & TMASS$^{\dagger*}$~\ct{wang2024text} & 43.4 & 70.6 & 81.0 & 41.8 & 71.9 & 81.4 & 56.1 & 88.1 & 93.8 & 628.1 \\
 & RAP~\ct{cao2024rap} & 42.6 & 70.4 & 79.6 & 44.9 & 73.7 & 83.1 & - & - & - & - \\
 & SHE-Net~\ct{yu2025she} & 45.6 & \textbf{75.6} & 83.6 & \underline{47.6} & \underline{76.8} & \underline{85.5} & - & - & - & - \\
 & GLSCL$^{*}$~\ct{zhang2025text} & 45.5 & \underline{75.4} & \underline{84.3} & - & - & - & - & - & - & - \\
 & TempMe$^{*}$~\ct{shentempme} & \textbf{47.2} & 73.7 & 81.3 & - & - & - & - & - & - & - \\
 & TV-ProxyNet$^{\dagger*}$~\ct{xiao2025text} & \underline{46.8} & 72.1 & 81.3 & 46.6 & 76.0 & 85.0 & 58.9 & \underline{89.1} & \textbf{94.4} & \underline{650.2} \\
 & Video-ColBERT$^{\dagger*}$~\ct{reddy2025video} & 40.2 & 68.6 & 79.3 & 43.9 & 73.9 & 83.5 & 57.1 & 87.4 & 93.1 & 627.0 \\
\rowcolor{gray!15}
\cellcolor{white} & \textbf{\shortname (Ours)} & \underline{46.8} & 74.0 & 83.5 & \textbf{47.9} & \textbf{77.7} & \textbf{85.9} & \underline{59.0} & \textbf{89.2} & 94.1 & \textbf{658.1} \\
\hline
\multirow{11}{*}{\rotatebox{90}{\textit{ViT-B/16}}} & CLIP4Clip$^{*}$~\ct{luo2022clip4clip} & 45.1 & 71.5 & 81.4 & 47.6 & 78.0 & 86.4 & 60.9 & 90.6 & 95.4 & 656.9 \\
 & XPool$^{\dagger*}$~\ct{gorti2022x} & 44.1 & 73.5 & 83.4 & 49.5 & 79.6 & 87.9 & 62.0 & 91.0 & 95.6 & 666.6 \\
 & TS2Net$^{*}$~\ct{liu2022ts2} & 43.1 & 72.3 & 81.5 & 47.0 & 78.0 & 86.6 & 62.1 & 91.2 & 95.8 & 657.6 \\
 & CLIP-VIP$^{\dagger*}$~\ct{xue2023clip} & 47.2 & 75.5 & 84.4 & 49.2 & 79.1 & 87.5 & \underline{63.8} & \underline{91.6} & \textbf{96.0} & 674.3 \\
 & ProST$^{*}$~\ct{li2023progressive} & 47.0 & 75.5 & 84.2 & 48.1 & 78.6 & 87.4 & 62.0 & 91.3 & 95.8 & 669.9 \\
 & TMASS$^{\dagger*}$~\ct{wang2024text} & 42.1 & 73.1 & 82.3 & 44.9 & 75.2 & 84.2 & 60.8 & 90.5 & 95.3 & 648.4 \\
 & GLSCL$^{*}$~\ct{zhang2025text} & 49.1 & \underline{76.5} & \underline{85.7} & - & - & - & - & - & - & - \\
 & TempMe$^{*}$~\ct{shentempme} & \underline{50.2} & 76.4 & 85.0 & - & - & - & - & - & - & - \\
 & TV-ProxyNet$^{\dagger*}$~\ct{xiao2025text} & 47.9 & 75.2 & 84.5 & \underline{49.7} & \underline{79.7} & \underline{88.0} & \textbf{64.0} & \textbf{91.7} & \underline{95.9} & \underline{676.6} \\
 & Video-ColBERT$^{\dagger*}$~\ct{reddy2025video} & 44.5 & 71.7 & 82.1 & 47.7 & 78.3 & 86.6 & 61.1 & 90.2 & 95.3 & 657.5 \\
\rowcolor{gray!15}
\cellcolor{white} & \textbf{\shortname (Ours)} & \textbf{51.5} & \textbf{78.7} & \textbf{86.8} & \textbf{50.9} & \textbf{80.7} & \textbf{88.3} & 63.6 & 91.5 & 95.7 & \textbf{687.7}
\end{tabular}

}

\vspace{-8pt}
\caption{Text-to-video results on DiDeMo, MSVD, and VATEX. 
Note that we have resolved the data-leakage problem in the original TMASS codes, 
and re-implemented XPool$^{\dagger}$, CLIP-VIP$^{\dagger}$, TMASS$^{\dagger}$, TV-ProxyNet$^{\dagger}$, and Video-ColBERT$^{\dagger}$ using CLIP4Clip codebase to align with other methods. 
$^{*}$ denotes the reproduced results, all of which are fine-tuned on ViT-B/32 or ViT-B/16. 
}
\label{tab:main_combine_t2v}

\vspace{-10pt}
\end{table*}

\textit{\textbf{Training \e.}}
In practice, \e is trained by $\mathcal{L}_{\text{\esup}} = -\mathbb{E}_{\{\vtext,\vframes\} \sim p(\vtext,\vframes)} [ \log p_{\theta}(\vtext,\vframes) ]$, 
this negative log-likelihood loss can be reformulated as: 
\begin{small}
\begin{equation}
    \begin{aligned}
        \mathcal{L}_{\text{\esup}} \approx \frac{1}{B}\sum_{i=1}^B E_{\theta}(\rvt_i,\vframes_i) 
        - \frac{1}{B}\sum_{j=1}^{B} E_{\theta} ( \Tilde{\vtext}_j,\Tilde{\vframes}_j ),
    \end{aligned}
\label{eq:eam_loss}
\end{equation}
\end{small}
where $\{(\Tilde{\vtext}_j,\Tilde{\vframes}_j)\}_{j=1}^{B}$ are fake text-video pairs sampled from $ p_\theta(\Tilde{\vtext}, \Tilde{\vframes}) $ using a $K$-step Markov chain Monte Carlo (MCMC) sampling:
\begin{small}
\begin{equation}
    \left\{
    \begin{aligned}
        \Tilde{\vtext}_j^{(k+1)} &= \Tilde{\vtext}_j^{(k)} - \eta \nabla_{\Tilde{\vtext}_j^{(k)}} E_\theta (\Tilde{\vtext}_j^{(k)},\Tilde{\vframes}_j^{(k)} ) + \bm{\eps}_{t_j}^{(k+1)}, \\
        \Tilde{\vframes}_j^{(k+1)} &= \Tilde{\vframes}_j^{(k)} - \eta \nabla_{\Tilde{\vframes}_j^{(k)}} E_\theta (\Tilde{\vtext}_j^{(k)},\Tilde{\vframes}_j^{(k)} ) + \bm{\eps}_{v_j}^{(k+1)}, \\
    \end{aligned}
    \right.
\label{eq:mcmc_sampling}
\end{equation}
\end{small}
where $\Tilde{\vtext}_j^{(0)} \in \mathbb{R}^{d}$ and $\Tilde{\vframes}_j^{(0)} \in \mathbb{R}^{M \times d}$ are given initial samples,  $\eta$ is the step size, and $\bm{\eps}^{(k)} \sim \mathcal{N} \left(0, \sigma^2 \right)$ with a given noise $\sigma^2$. 
We stop the gradients of $\Tilde{\vtext}$ and $\Tilde{\vframes}$ after sampling. 
More details of the derivation and sampling algorithm can be found in~\appcref{sec:app.ebm}. 
\e is only involved in the training process, 
and thus will not increase inference cost. 

We incorporate $\loss_{\text{eam}}$ to define the total objective: 
\begin{small}
\begin{equation}
\begin{aligned}
\loss_{\text{total}} =& \loss_{\text{match}} + \lambda_{\text{\esup}} \loss_{\text{\esup}} \\
=& \loss_{\text{ce}}(\rvt^{\text{gen}},\rvv) + \lambda_{\text{sup}} \loss_{\text{ce}}(\rvt^{\text{sup}},\rvv) + \lambda_{\text{\esup}} \loss_{\text{\esup}}, 
\end{aligned}
\label{eq:total_loss_1}
\end{equation}
\end{small}
where $\lambda_{\text{\esup}}$ is a coefficient. 

\subsubsection{Sigmoid Loss}\label{sec:sigloss}

The softmax-based contrastive loss ($\loss_{\text{ce}}$) is sensitive to negative samples and batch size due to the batch normalization~\ct{chuang2020debiased,karpukhin2020dense,zhai2023sigmoid}. 
Instead, the sigmoid loss~\ct{zhai2023sigmoid} models each text-video pair independently, enabling more effective alignment and stable optimization, so is naturally suitable for the multi-matched scenario of TVR, where one text may be semantically similar to multiple videos or vice versa. 
\begin{small}
\begin{equation}
    \loss_{\text{sig}}(\rvt,\rvv) = -\frac{1}{B}\sum_{i=1}^B\sum_{j=1}^B \log \frac{1}{1+e^{
    \mathbb{I}_{ij}\left(\tau \cdot \operatorname{s}(\rvt_i, \rvv_j) + b\right)}} 
\label{eq:sig_loss}
\end{equation}
\end{small}
where $\mathbb{I}_{ij}$ is a indicator for positive pairs ($+1$) and negative pairs ($-1$), $\tau=e^{\tau_p}$, $\tau_p$ and $b$ are learnable and initialized as $\tau_p=4.77$ and $b=-12.93$, according to the pre-training in~\ct{zhai2023sigmoid}. 
Finally, we replace $\loss_{\text{ce}}$ with $\loss_{\text{sig}}$ in \cref{eq:total_loss_1}:
\begin{small}
\begin{equation}
    \loss_{\text{total}} = \loss_{\text{sig}}(\rvt^{\text{gen}},\rvv) + \lambda_{\text{sup}} \loss_{\text{sig}}(\rvt^{\text{sup}},\rvv) + \lambda_{\text{\esup}} \loss_{\text{\esup}}.
\label{eq:total_loss_2}
\end{equation}
\end{small}

\vspace{-7mm}
\section{Experiment}

\subsection{Experimental Settings}

\textbf{\textit{Datasets and Metrics.}}
We use four benchmark datasets: \textbf{MSRVTT}~\ct{xu2016msr}, \textbf{DiDeMo}~\ct{anne2017localizing}, \textbf{MSVD}~\ct{chen2011collecting}, and \textbf{VATEX}~\ct{wang2019vatex}. These datasets contain sufficient video clips, each of which has multiple captions. 
We use the standard TVR metrics~\ct{luo2022clip4clip}: Recall (\%) at rank 1/5/10 (R@1/5/10), Median Rank (MdR), Mean Rank (MnR), and Rsum (R@1$+$R@5$+$R@10); and also the total test time (sec.). 
More details can be found in \appcref{sec:app.exp_settings}.

\textbf{\textit{Implementation.}}
We initialized our model by CLIP pre-trained models (ViT-B/32 and ViT-B/16)~\ct{radford2021learning}, so hidden dimension $d=512$.
We uniformly sample $M=12$ frames per video ($64$ for DiDeMo) and the max length of text is $32$ ($64$ for DiDeMo). 
In \g, we sample $S=20$ text candidates. 
For MCMC sampling in \e, step number $K=20$, step size $\eta=1$, and noise variance $\sigma^2=0.005$. 
For loss coefficient, we set $\lambda_{\text{sup}}=0.8$ following~\cite{wang2024text} and $\lambda_{\text{eam}}=1.0$ by default. 
We train on two A100 GPUs for $5$ epochs with batch size $B=64$, weight decay $0.2$, dropout $0.3$, learning rate $10^{-7}$ for CLIP modules and $10^{-4}$ for non-CLIP modules; and we evaluate on one A100 with batch size $8$. 
More details can be found in \appcref{sec:app.exp_settings}.

\subsection{Comparison to State-of-the-art Models}
\label{sec:compare_sota}

We compare our proposed \shortname with existing state-of-the-art (SOTA) models~\ct{luo2022clip4clip,gorti2022x,liu2022ts2,xue2023clip,wang2023unified,li2023progressive,wang2024text,zhang2025text,shentempme,xiao2025text,reddy2025video}. 
We show the performance in~\cref{tab:main_msrvtt9k,tab:main_combine_t2v}, and the test time in~\cref{fig:r1-time}. 
See~\appcref{sec:app.additional_analysis} for further analysis and complexity details.
Note that we have addressed the data-leakage problem in the original TMASS codes, and re-implemented XPool$^{\dagger}$, CLIP-VIP$^{\dagger}$, TMASS$^{\dagger}$, TV-ProxyNet$^{\dagger}$, and Video-ColBERT$^{\dagger}$ using CLIP4Clip codebase to align with other models for fair comparisons. 

Overall, \shortname achieves the best text-to-video R@1 performance across datasets, 
outperforming the second-best by 2.00\% on MSRVTT, 2.59\% on DiDeMo, and 2.41\% on MSVD, respectively. 
And our model consistently reaches the highest Rsum results, demonstrating the superiority and universality of our proposed model.

\begin{figure}
    \centering
    \includegraphics[width=0.9\linewidth]{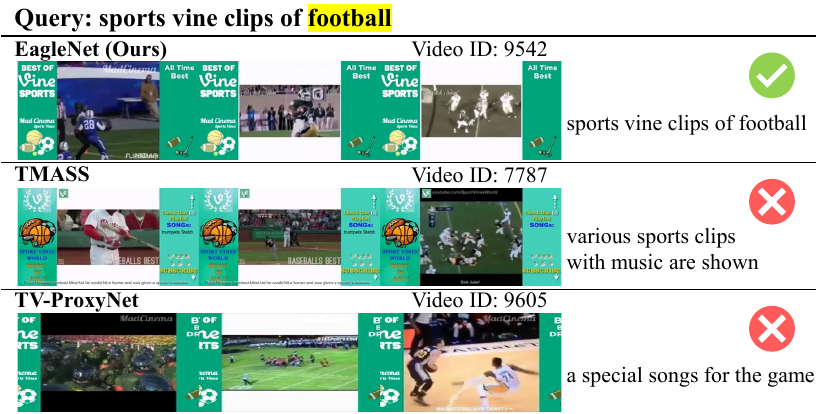}
    \vspace{-8pt}
    \caption{Visualization of \shortname, TMASS, and TV-ProxyNet.}
    \label{fig:visualization-main}
    \vspace{-15pt}
\end{figure}

\begin{table*}[!t]
\centering

\resizebox{0.7\linewidth}{!}{

\begin{tabular}{c|ccc|ccc|ccc|ccc}
 & \multicolumn{3}{c|}{\textbf{Components}} & \multicolumn{3}{c|}{\textbf{MSRVTT}} & \multicolumn{3}{c|}{\textbf{DiDeMo}} & \multicolumn{3}{c}{\textbf{Avg}} \\
 & \g & \e & $\loss_{\text{sig}}$ & \textbf{R@1↑} & \textbf{R@5↑} & \textbf{R@10↑} & \textbf{R@1↑} & \textbf{R@5↑} & \textbf{R@10↑} & \textbf{R@1↑} & \textbf{R@5↑} & \textbf{R@10↑} \\
\hline
 &  &  &  & 48.5 & 74.1 & 83.4 & 42.1 & 73.1 & 82.3 & 45.3 & 73.6 & 82.9 \\
 & \checkmark &  &  & 48.8 & 75.6 & 83.9 & 47.9 & 75.2 & 84.1 & 48.4 & 75.4 & 84.0 \\
 &  & \checkmark &  & 49.0 & 75.3 & 84.0 & 43.4 & 73.6 & 82.7 & 46.2 & 74.5 & 83.4 \\
 & \checkmark & \checkmark &  & \underline{50.5} & \underline{76.5} & 84.2 & \underline{49.2} & 76.6 & \underline{85.3} & \underline{49.9} & 76.6 & \underline{84.8} \\
\hline
 &  &  & \checkmark & 47.8 & 75.9 & \underline{84.4} & 43.9 & 75.6 & 83.6 & 45.9 & 75.8 & 84.0 \\
 & \checkmark &  & \checkmark & 49.9 & \textbf{77.3} & 84.3 & 49.1 & \underline{77.5} & 85.0 & 49.5 & \underline{77.4} & 84.7 \\
 &  & \checkmark & \checkmark & 48.6 & 75.3 & 83.7 & 43.8 & 75.6 & 83.8 & 46.2 & 75.5 & 83.8 \\
\rowcolor{gray!15}
\textbf{\shortname} & \checkmark & \checkmark & \checkmark & \textbf{51.0} & 76.2 & \textbf{85.6} & \textbf{51.5} & \textbf{78.7} & \textbf{86.8} & \textbf{51.3} & \textbf{77.5} & \textbf{86.2}
\end{tabular}

}

\vspace{-8pt}
\caption{Text-to-video results on different components.}
\label{tab:ablaion-components}

\vspace{-15pt}
\end{table*}
\begin{table}[!t]
\centering

\resizebox{0.9\linewidth}{!}{

\begin{tabular}{l|ccc|ccc}
\multicolumn{1}{c|}{\textbf{Energy}} & \multicolumn{3}{c|}{\textbf{MSRVTT}} & \multicolumn{3}{c}{\textbf{DiDeMo}} \\
\multicolumn{1}{c|}{\textbf{Function}} & \textbf{R@1↑} & \textbf{R@5↑} & \textbf{R@10↑} & \textbf{R@1↑} & \textbf{R@5↑} & \textbf{R@10↑} \\
\hline
\texttt{CosSim} & 50.0 & \textbf{76.9} & \underline{85.5} & 50.2 & \underline{78.3} & 86.3 \\
\rowcolor{gray!15}
\texttt{Bilinear} & \textbf{51.0} & 76.2 & \textbf{85.6} & \underline{51.3} & \underline{78.3} & \underline{86.6} \\
\texttt{MLP} & \underline{50.3} & \underline{76.6} & 85.1 & \textbf{51.5} & \textbf{78.7} & \textbf{86.8}
\end{tabular}

}

\vspace{-8pt}
\caption{Text-to-video on different energy functions.}
\label{tab:ablaion-efunc}

\vspace{-5pt}
\end{table}
\begin{table}[!t]
\centering

\resizebox{0.9\linewidth}{!}{

\begin{tabular}{c|ccc|ccc}
\textbf{Pooling} & \multicolumn{3}{c|}{\textbf{MSRVTT}} & \multicolumn{3}{c}{\textbf{DiDeMo}} \\
\textbf{Form} & \textbf{R@1↑} & \textbf{R@5↑} & \textbf{R@10↑} & \textbf{R@1↑} & \textbf{R@5↑} & \textbf{R@10↑} \\
\hline
w/o \e & 49.9 & \textbf{77.3} & 84.3 & 49.1 & 77.5 & 85.0 \\
$E_{\theta}(\rvt,\rvv)$ & 48.8 & 76.3 & 84.6 & 47.8 & 77.1 & 85.2 \\
\texttt{Maxpool} & 50.4 & 76.5 & \underline{85.5} & \underline{51.3} & \textbf{79.0} & \underline{86.6} \\
\texttt{Minpool} & \underline{50.5} & \underline{77.2} & 84.9 & 50.9 & 78.5 & 86.5 \\
\rowcolor{gray!15}
\texttt{Avgpool} & \textbf{51.0} & 76.2 & \textbf{85.6} & \textbf{51.5} & \underline{78.7} & \textbf{86.8}
\end{tabular}

}

\vspace{-8pt}
\caption{Text-to-video on different pooling approach for $E_{\theta}$.}
\label{tab:ablaion-epool}

\vspace{-10pt}
\end{table}

Both TMASS~\ct{wang2024text} and TV-ProxyNet~\ct{xiao2025text} focus solely on the interactions between text and frames/video, so they fail to exploit temporal information within the video, resulting in suboptimal performance (MSRVTT text-to-video R@1 48.5\% for TMASS and 49.5\% for TV-ProxyNet). 
From~\cref{fig:visualization-main}, they might retrieve incorrect videos with similar semantics but different contexts. 
In contrast, \shortname also leverages the internal frame-to-frame interactions within the video when expanding text semantics. 
Modeling such frame-level relationships enables the enriched text embedding to capture both the global video semantics and the temporal contextual information, leading to SOTA performance across benchmarks. 
Visualization of not considering frame context (without \g) is presented in \appcref{sec:app.visualization-components}. 
Moreover, TMASS, based on XPool~\ct{gorti2022x}, designs a stochastic text modeling strategy to sample text embeddings with richer semantics. 
However, after addressing the data-leakage issue, its text-to-video R@1 result (48.5\%) is worse than that of XPool (49.2\%) on MSRVTT, underperforms the runner-up by 16.14\% on DiDeMo, and achieves unsatisfactory performance on MSVD (44.9\%) and VATEX (60.8\%). 
These results suggest that stochastically sampling a single text embedding per text-video pair may introduce undesired noise, degrading the semantic quality of the final text representation. \shortname instead extracts frame-relevant information from multiple text candidates, producing an accurate text embedding. 
We provide detailed visualization comparisons of different models in~\appcref{sec:app.visualization-models}.

\subsection{Ablation Study}

\subsubsection{The Effect of Different Components}

From~\cref{tab:ablaion-components}, we first analyze the effect of different components (\g and \e). 
The base model (Row 1, R@1 48.5\% on MSRVTT and 42.1\% on DiDeMo) only uses the interactions between text and frames, yielding worse results than the baseline, XPool (49.2\% on MSRVTT and 44.1\% on DiDeMo). 
In contrast, \g (Row 2) incorporates frame-frame relationships and thus captures the frame contextual information, improving the final performance (R@1 48.8\% on MSRVTT and 47.9\% on DiDeMo). 
Moreover, as mentioned in~\cref{sec:compare_sota}, the stochastic text modeling of TMASS (Row 1) may bring undesired noise, thereby also lowering the performance compared to its baseline, XPool. 
\g instead extracts frame-related information from multiple text candidates, resulting in an accurate enriched text embedding. 
Meanwhile, \e models the distribution of real text-video pairs from a fine-grained perspective, enhancing the ability of the model to consider fine-grained matching relevance and consequently improving the performance (Row 3, 49.0\% on average). 
Combining them, \e can enhance \g's performance (Row 4, 49.9\% on average) in learning relationships by utilizing the fine-grained information in text-frame interactions. 

Moreover, we observe an overall improvement when using $\loss_{\text{sig}}$ over $\loss_{\text{ce}}$ (Row 5 to 8). 
Since sigmoid loss models each text-video pair independently, it is unaffected by the batch size and negative samples, enabling more effective alignment and stable optimization. 
These features make it more suitable for TVR scenarios. 

Detailed visualization comparison of different components can be found in~\appcref{sec:app.visualization-components}. 

\subsubsection{The Selection of Energy Function}

We compare the performance of different energy functions described in~\cref{sec:eam} (\cref{tab:ablaion-efunc}). 
On the dataset MSRVTT, the \texttt{Bilinear} shows the best R@1 results (51.0\%).
On DiDeMo, the best result (51.5\%) is given by \texttt{MLP}. 
This indicates that we need to select a suitable one for different applications. 

Although \texttt{CosSim} is learning-free and shows comparable performance, such a non-parametric energy function limits the capability of dealing with complicated scenarios. 
The \texttt{Bilinear} can be regarded as \texttt{CosSim} with learnable parameters and is more flexible. 
\texttt{MLP} adaptively learn the connection between text and video and is no longer limited to a cosine similarity form, 
so it can discover more implicit information in some cases but has more leanable parameters. 
Overall, we choose \texttt{Bilinear} as our final implementation. 

\subsubsection{The Pooling Approach for Text-Frame Energy}\label{sec:pooling_energy}

Here we discuss the pooling approach for energy function, \texttt{Avgpool} ($\frac{1}{M}\sum_{i=1}^M E_{\theta}(\rvt,\rvf_i)$), \texttt{Maxpool} ($\max_i E_{\theta}(\rvt,\rvf_i)$), and \texttt{Minpool} ($\min_i E_{\theta}(\rvt,\rvf_i)$). 
Since the energy of a matched pair is low, \texttt{Minpool} selects the frame that is most relevant to the given text. 
On the contrary, \texttt{Maxpool} selects the most irrelevant frame, which can be seen as a lower bound of text-video relevance. 
Besides, we can also use the global video embedding to calculate the text-video energy ($E_{\theta}(\rvt,\rvv)$). 

From~\cref{tab:ablaion-epool}, 
\texttt{Avgpool} shows the best R@1 performance on both MSRVTT (51.0\%) and DiDeMo (51.5\%), 
suggesting that considering all the text-frame interactions brings the most benefit to the retrieval. 
Regarding the other two approaches, selecting the most relevant frame yields better results on MSRVTT (\texttt{Minpool}, 50.5\%); but improving the lower bound to enhance the overall relevance is more suitable for DiDeMo (\texttt{Maxpool}, 51.3\%). 

Besides, fusing frames before calculating energy shows lower performance (48.8\% and 47.8\%) than not using \e (49.9\% and 49.1\%). 
Therefore, learning from fine-grained text-frame interactions can better improve the model in identifying the actual text-video pair. 
The fused video embedding provides only a global view, missing fine-grained details.

\section{Conclusions}
\vspace{-2mm}

In this paper, we present \shortname, a novel text-video retrieval framework that generates accurate and context-aware enriched text embeddings by exploring fine-grained relationships among text and video frames. 
We further employ an energy-based model to capture detailed interactions, which facilitates learning of relationships and enhances matching performance from a fine-grained perspective. 
Moreover, we instead use sigmoid loss for more effective alignment and stable training. 
Extensive experiments across four datasets demonstrate the superior performance of \shortname compared with other state-of-the-art methods. 
\clearpage
\section*{Acknowledgements}

This work was supported in part 
by the Shenzhen Science and Technology Program under Grant No.~KQTD20221101093559018, 
Grant No.~CJGJZD20240729141505007, 
and Grant No.~SYSRD20250529113401002, 
by the National Natural Science Foundation of China (NSFC) under Grant No. 62441619 and Grant No.~62411540034. 

{
    \small
    \bibliographystyle{ieeenat_fullname}
    \bibliography{main}

\begin{thebibliography}{59}
\providecommand{\natexlab}[1]{#1}
\providecommand{\url}[1]{\texttt{#1}}
\expandafter\ifx\csname urlstyle\endcsname\relax
  \providecommand{\doi}[1]{doi: #1}\else
  \providecommand{\doi}{doi: \begingroup \urlstyle{rm}\Url}\fi

\bibitem[Anne~Hendricks et~al.(2017)Anne~Hendricks, Wang, Shechtman, Sivic, Darrell, and Russell]{anne2017localizing}
Lisa Anne~Hendricks, Oliver Wang, Eli Shechtman, Josef Sivic, Trevor Darrell, and Bryan Russell.
\newblock Localizing moments in video with natural language.
\newblock In \emph{Proceedings of the IEEE/CVF International Conference on Computer Vision}, pages 5803--5812, 2017.

\bibitem[Bain et~al.(2021)Bain, Nagrani, Varol, and Zisserman]{bain2021frozen}
Max Bain, Arsha Nagrani, G{\"u}l Varol, and Andrew Zisserman.
\newblock Frozen in time: A joint video and image encoder for end-to-end retrieval.
\newblock In \emph{Proceedings of the IEEE/CVF International Conference on Computer Vision}, pages 1728--1738, 2021.

\bibitem[Cao et~al.(2024)Cao, Tang, Huang, Jin, Zhang, Liu, Chen, Liang, Yuan, and Li]{cao2024rap}
Meng Cao, Haoran Tang, Jinfa Huang, Peng Jin, Can Zhang, Ruyang Liu, Long Chen, Xiaodan Liang, Li Yuan, and Ge Li.
\newblock Rap: Efficient text-video retrieval with sparse-and-correlated adapter.
\newblock In \emph{Findings of the Association for Computational Linguistics}, pages 7160--7174, 2024.

\bibitem[Chen and Dolan(2011)]{chen2011collecting}
David Chen and William~B Dolan.
\newblock Collecting highly parallel data for paraphrase evaluation.
\newblock In \emph{Proceedings of the Annual Meeting of the Association for Computational Linguistics: Human Language Technologies}, pages 190--200, 2011.

\bibitem[Chen et~al.(2020)Chen, Zhao, Jin, and Wu]{chen2020fine}
Shizhe Chen, Yida Zhao, Qin Jin, and Qi Wu.
\newblock Fine-grained video-text retrieval with hierarchical graph reasoning.
\newblock In \emph{Proceedings of the IEEE/CVF Conference on Computer Vision and Pattern Recognition}, pages 10638--10647, 2020.

\bibitem[Chen et~al.(2025)Chen, Luo, Song, Dai, Tang, and Cao]{chendecoupled}
Yuhan Chen, Yihong Luo, Yifan Song, Pengwen Dai, Jing Tang, and Xiaochun Cao.
\newblock Decoupled graph energy-based model for node out-of-distribution detection on heterophilic graphs.
\newblock In \emph{International Conference on Learning Representations}, 2025.

\bibitem[Cheng et~al.(2021)Cheng, Lin, Wu, Yang, and Shen]{cheng2021improving}
Xing Cheng, Hezheng Lin, Xiangyu Wu, Fan Yang, and Dong Shen.
\newblock Improving video-text retrieval by multi-stream corpus alignment and dual softmax loss.
\newblock \emph{arXiv preprint arXiv:2109.04290}, 2021.

\bibitem[Chuang et~al.(2020)Chuang, Robinson, Lin, Torralba, and Jegelka]{chuang2020debiased}
Ching-Yao Chuang, Joshua Robinson, Yen-Chen Lin, Antonio Torralba, and Stefanie Jegelka.
\newblock Debiased contrastive learning.
\newblock \emph{Advances in Neural Information Processing Systems}, 33:\penalty0 8765--8775, 2020.

\bibitem[Croitoru et~al.(2021)Croitoru, Bogolin, Leordeanu, Jin, Zisserman, Albanie, and Liu]{croitoru2021teachtext}
Ioana Croitoru, Simion-Vlad Bogolin, Marius Leordeanu, Hailin Jin, Andrew Zisserman, Samuel Albanie, and Yang Liu.
\newblock Teachtext: Crossmodal generalized distillation for text-video retrieval.
\newblock In \emph{Proceedings of the IEEE/CVF International Conference on Computer Vision}, pages 11583--11593, 2021.

\bibitem[Dong et~al.(2021)Dong, Li, Xu, Yang, Yang, Wang, and Wang]{dong2021dual}
Jianfeng Dong, Xirong Li, Chaoxi Xu, Xun Yang, Gang Yang, Xun Wang, and Meng Wang.
\newblock Dual encoding for video retrieval by text.
\newblock \emph{IEEE Transactions on Pattern Analysis and Machine Intelligence}, 44\penalty0 (8):\penalty0 4065--4080, 2021.

\bibitem[Du and Mordatch(2019)]{du2019implicit}
Yilun Du and Igor Mordatch.
\newblock Implicit generation and modeling with energy based models.
\newblock \emph{Advances in Neural Information Processing Systems}, 32:\penalty0 3608--3618, 2019.

\bibitem[Fang et~al.(2021)Fang, Xiong, Xu, and Chen]{fang2021clip2video}
Han Fang, Pengfei Xiong, Luhui Xu, and Yu Chen.
\newblock Clip2video: Mastering video-text retrieval via image clip.
\newblock \emph{arXiv preprint arXiv:2106.11097}, 2021.

\bibitem[Fang et~al.(2022)Fang, Wang, Zhuo, Huang, Ma, Wei, and Wei]{fang2022concept}
Sheng Fang, Shuhui Wang, Junbao Zhuo, Qingming Huang, Bin Ma, Xiaoming Wei, and Xiaolin Wei.
\newblock Concept propagation via attentional knowledge graph reasoning for video-text retrieval.
\newblock In \emph{Proceedings of the ACM International Conference on Multimedia}, pages 4789--4800, 2022.

\bibitem[Gabeur et~al.(2020)Gabeur, Sun, Alahari, and Schmid]{gabeur2020multi}
Valentin Gabeur, Chen Sun, Karteek Alahari, and Cordelia Schmid.
\newblock Multi-modal transformer for video retrieval.
\newblock In \emph{Proceedings of European Conference on Computer Vision}, pages 214--229, 2020.

\bibitem[Gao et~al.(2023)Gao, Xiong, Gao, Jia, Pan, Bi, Dai, Sun, Wang, and Wang]{gao2023retrieval}
Yunfan Gao, Yun Xiong, Xinyu Gao, Kangxiang Jia, Jinliu Pan, Yuxi Bi, Yi Dai, Jiawei Sun, Meng Wang, and Haofen Wang.
\newblock Retrieval-augmented generation for large language models: A survey.
\newblock \emph{arXiv preprint arXiv:2312.10997}, 2023.

\bibitem[Ge et~al.(2022)Ge, Ge, Liu, Li, Shan, Qie, and Luo]{ge2022bridging}
Yuying Ge, Yixiao Ge, Xihui Liu, Dian Li, Ying Shan, Xiaohu Qie, and Ping Luo.
\newblock Bridging video-text retrieval with multiple choice questions.
\newblock In \emph{Proceedings of the IEEE/CVF Conference on Computer Vision and Pattern Recognition}, pages 16167--16176, 2022.

\bibitem[Gorti et~al.(2022)Gorti, Vouitsis, Ma, Golestan, Volkovs, Garg, and Yu]{gorti2022x}
Satya~Krishna Gorti, No{\"e}l Vouitsis, Junwei Ma, Keyvan Golestan, Maksims Volkovs, Animesh Garg, and Guangwei Yu.
\newblock X-pool: Cross-modal language-video attention for text-video retrieval.
\newblock In \emph{Proceedings of the IEEE/CVF Conference on Computer Vision and Pattern Recognition}, pages 5006--5015, 2022.

\bibitem[Hanu et~al.(2022)Hanu, Thewlis, Asano, and Rupprecht]{hanu2022vtc}
Laura Hanu, James Thewlis, Yuki~M Asano, and Christian Rupprecht.
\newblock Vtc: Improving video-text retrieval with user comments.
\newblock In \emph{Proceedings of European Conference on Computer Vision}, pages 616--633, 2022.

\bibitem[Hao et~al.(2021)Hao, Zhou, Wu, Zhang, Li, and Wang]{hao2021multi}
Xiaoshuai Hao, Yucan Zhou, Dayan Wu, Wanqian Zhang, Bo Li, and Weiping Wang.
\newblock Multi-feature graph attention network for cross-modal video-text retrieval.
\newblock In \emph{Proceedings of the International Conference on Multimedia Retrieval}, pages 135--143, 2021.

\bibitem[He et~al.(2016)He, Zhang, Ren, and Sun]{he2016deep}
Kaiming He, Xiangyu Zhang, Shaoqing Ren, and Jian Sun.
\newblock Deep residual learning for image recognition.
\newblock In \emph{Proceedings of the IEEE/CVF Conference on Computer Vision and Pattern Recognition}, pages 770--778, 2016.

\bibitem[Hu et~al.(2022)Hu, Chen, Wang, Zhou, Dong, and Li]{hu2022lightweight}
Fan Hu, Aozhu Chen, Ziyue Wang, Fangming Zhou, Jianfeng Dong, and Xirong Li.
\newblock Lightweight attentional feature fusion: A new baseline for text-to-video retrieval.
\newblock In \emph{Proceedings of European Conference on Computer Vision}, pages 444--461, 2022.

\bibitem[Jin et~al.(2022)Jin, Huang, Liu, Wu, Ge, Song, Clifton, and Chen]{jin2022expectation}
Peng Jin, Jinfa Huang, Fenglin Liu, Xian Wu, Shen Ge, Guoli Song, David Clifton, and Jie Chen.
\newblock Expectation-maximization contrastive learning for compact video-and-language representations.
\newblock \emph{Advances in Neural Information Processing Systems}, 35:\penalty0 30291--30306, 2022.

\bibitem[Jin et~al.(2021)Jin, Zhao, Zhang, Zhu, He, and Zhuang]{jin2021hierarchical}
Weike Jin, Zhou Zhao, Pengcheng Zhang, Jieming Zhu, Xiuqiang He, and Yueting Zhuang.
\newblock Hierarchical cross-modal graph consistency learning for video-text retrieval.
\newblock In \emph{Proceedings of the International ACM SIGIR Conference on Research and Development in Information Retrieval}, pages 1114--1124, 2021.

\bibitem[Karpukhin et~al.(2020)Karpukhin, O{\u{g}}uz, Min, Lewis, Wu, Edunov, Chen, and Yih]{karpukhin2020dense}
Vladimir Karpukhin, Barlas O{\u{g}}uz, Sewon Min, Patrick Lewis, Ledell Wu, Sergey Edunov, Danqi Chen, and Wen~Tau Yih.
\newblock Dense passage retrieval for open-domain question answering.
\newblock In \emph{Conference on Empirical Methods in Natural Language Processing}, pages 6769--6781. Association for Computational Linguistics (ACL), 2020.

\bibitem[Kipf and Welling(2017)]{kipf2017semi}
Thomas~N Kipf and Max Welling.
\newblock Semi-supervised classification with graph convolutional networks.
\newblock In \emph{International Conference on Learning Representations}, 2017.

\bibitem[Lei et~al.(2021)Lei, Li, Zhou, Gan, Berg, Bansal, and Liu]{lei2021less}
Jie Lei, Linjie Li, Luowei Zhou, Zhe Gan, Tamara~L Berg, Mohit Bansal, and Jingjing Liu.
\newblock Less is more: Clipbert for video-and-language learning via sparse sampling.
\newblock In \emph{Proceedings of the IEEE/CVF Conference on Computer Vision and Pattern Recognition}, pages 7331--7341, 2021.

\bibitem[Li et~al.(2023)Li, Xie, Zhao, Xie, Ge, Zheng, Zhao, and Zhang]{li2023progressive}
Pandeng Li, Chen-Wei Xie, Liming Zhao, Hongtao Xie, Jiannan Ge, Yun Zheng, Deli Zhao, and Yongdong Zhang.
\newblock Progressive spatio-temporal prototype matching for text-video retrieval.
\newblock In \emph{Proceedings of the IEEE/CVF International Conference on Computer Vision}, pages 4100--4110, 2023.

\bibitem[Lin et~al.(2022)Lin, Wu, Liang, Zhang, Ge, Zheng, and Shen]{lin2022text}
Chengzhi Lin, Ancong Wu, Junwei Liang, Jun Zhang, Wenhang Ge, Wei-Shi Zheng, and Chunhua Shen.
\newblock Text-adaptive multiple visual prototype matching for video-text retrieval.
\newblock \emph{Advances in Neural Information Processing Systems}, 35:\penalty0 38655--38666, 2022.

\bibitem[Liu et~al.(2021)Liu, Fan, Qian, Chen, Ding, and Wang]{liu2021hit}
Song Liu, Haoqi Fan, Shengsheng Qian, Yiru Chen, Wenkui Ding, and Zhongyuan Wang.
\newblock Hit: Hierarchical transformer with momentum contrast for video-text retrieval.
\newblock In \emph{Proceedings of the IEEE/CVF International Conference on Computer Vision}, pages 11915--11925, 2021.

\bibitem[Liu et~al.(2020)Liu, Wang, Owens, and Li]{liu2020energy}
Weitang Liu, Xiaoyun Wang, John Owens, and Yixuan Li.
\newblock Energy-based out-of-distribution detection.
\newblock \emph{Advances in Neural Information Processing Systems}, 33:\penalty0 21464--21475, 2020.

\bibitem[Liu et~al.(2019)Liu, Albanie, Nagrani, and Zisserman]{liu2019use}
Yang Liu, Samuel Albanie, Arsha Nagrani, and Andrew Zisserman.
\newblock Use what you have: Video retrieval using representations from collaborative experts.
\newblock \emph{arXiv preprint arXiv:1907.13487}, 2019.

\bibitem[Liu et~al.(2022)Liu, Xiong, Xu, Cao, and Jin]{liu2022ts2}
Yuqi Liu, Pengfei Xiong, Luhui Xu, Shengming Cao, and Qin Jin.
\newblock Ts2-net: Token shift and selection transformer for text-video retrieval.
\newblock In \emph{Proceedings of European Conference on Computer Vision}, pages 319--335, 2022.

\bibitem[Luo et~al.(2022)Luo, Ji, Zhong, Chen, Lei, Duan, and Li]{luo2022clip4clip}
Huaishao Luo, Lei Ji, Ming Zhong, Yang Chen, Wen Lei, Nan Duan, and Tianrui Li.
\newblock Clip4clip: An empirical study of clip for end to end video clip retrieval and captioning.
\newblock \emph{Neurocomputing}, 508:\penalty0 293--304, 2022.

\bibitem[Ma et~al.(2022)Ma, Xu, Sun, Yan, Zhang, and Ji]{ma2022x}
Yiwei Ma, Guohai Xu, Xiaoshuai Sun, Ming Yan, Ji Zhang, and Rongrong Ji.
\newblock X-clip: End-to-end multi-grained contrastive learning for video-text retrieval.
\newblock In \emph{Proceedings of the ACM International Conference on Multimedia}, pages 638--647, 2022.

\bibitem[Oord et~al.(2018)Oord, Li, and Vinyals]{oord2018representation}
Aaron van~den Oord, Yazhe Li, and Oriol Vinyals.
\newblock Representation learning with contrastive predictive coding.
\newblock \emph{arXiv preprint arXiv:1807.03748}, 2018.

\bibitem[Patrick et~al.(2021)Patrick, Huang, Asano, Metze, Hauptmann, Henriques, and Vedaldi]{patricksupport}
Mandela Patrick, Po-Yao Huang, Yuki Asano, Florian Metze, Alexander~G Hauptmann, Joao~F Henriques, and Andrea Vedaldi.
\newblock Support-set bottlenecks for video-text representation learning.
\newblock In \emph{International Conference on Learning Representations}, 2021.

\bibitem[Radford et~al.(2021)Radford, Kim, Hallacy, Ramesh, Goh, Agarwal, Sastry, Askell, Mishkin, Clark, et~al.]{radford2021learning}
Alec Radford, Jong~Wook Kim, Chris Hallacy, Aditya Ramesh, Gabriel Goh, Sandhini Agarwal, Girish Sastry, Amanda Askell, Pamela Mishkin, Jack Clark, et~al.
\newblock Learning transferable visual models from natural language supervision.
\newblock In \emph{International Conference on Machine Learning}, pages 8748--8763, 2021.

\bibitem[Reddy et~al.(2025)Reddy, Martin, Yang, Yates, Sanders, Murray, Kriz, de~Melo, Van~Durme, and Chellappa]{reddy2025video}
Arun Reddy, Alexander Martin, Eugene Yang, Andrew Yates, Kate Sanders, Kenton Murray, Reno Kriz, Celso~M de Melo, Benjamin Van~Durme, and Rama Chellappa.
\newblock Video-colbert: Contextualized late interaction for text-to-video retrieval.
\newblock In \emph{Proceedings of the IEEE/CVF Conference on Computer Vision and Pattern Recognition}, pages 19691--19701, 2025.

\bibitem[Schlichtkrull et~al.(2018)Schlichtkrull, Kipf, Bloem, Van Den~Berg, Titov, and Welling]{schlichtkrull2018modeling}
Michael Schlichtkrull, Thomas~N Kipf, Peter Bloem, Rianne Van Den~Berg, Ivan Titov, and Max Welling.
\newblock Modeling relational data with graph convolutional networks.
\newblock In \emph{European Semantic Web Conference}, pages 593--607, 2018.

\bibitem[Shen et~al.(2025{\natexlab{a}})Shen, Hao, He, Zhang, Liu, Zhao, Han, and Ding]{shen2025temporal}
Leqi Shen, Tianxiang Hao, Tao He, Yifeng Zhang, Pengzhang Liu, Sicheng Zhao, Jungong Han, and Guiguang Ding.
\newblock Temporal modeling with frozen vision--language foundation models for parameter-efficient text--video retrieval.
\newblock \emph{IEEE Transactions on Neural Networks and Learning Systems}, 2025{\natexlab{a}}.

\bibitem[Shen et~al.(2025{\natexlab{b}})Shen, Hao, He, Zhao, Zhang, Bao, Ding, et~al.]{shentempme}
Leqi Shen, Tianxiang Hao, Tao He, Sicheng Zhao, Yifeng Zhang, Yongjun Bao, Guiguang Ding, et~al.
\newblock Tempme: Video temporal token merging for efficient text-video retrieval.
\newblock In \emph{International Conference on Learning Representations}, 2025{\natexlab{b}}.

\bibitem[Veli{\v{c}}kovi{\'c} et~al.(2018)Veli{\v{c}}kovi{\'c}, Cucurull, Casanova, Romero, Li{\`o}, and Bengio]{velivckovic2018graph}
Petar Veli{\v{c}}kovi{\'c}, Guillem Cucurull, Arantxa Casanova, Adriana Romero, Pietro Li{\`o}, and Yoshua Bengio.
\newblock Graph attention networks.
\newblock In \emph{International Conference on Learning Representations}, 2018.

\bibitem[Wang et~al.(2024)Wang, Sun, Wang, Liu, Dianat, Rabbani, Rao, and Tao]{wang2024text}
Jiamian Wang, Guohao Sun, Pichao Wang, Dongfang Liu, Sohail Dianat, Majid Rabbani, Raghuveer Rao, and Zhiqiang Tao.
\newblock Text is mass: Modeling as stochastic embedding for text-video retrieval.
\newblock In \emph{Proceedings of the IEEE/CVF Conference on Computer Vision and Pattern Recognition}, pages 16551--16560, 2024.

\bibitem[Wang et~al.(2020)Wang, Gao, Yang, and Xu]{wang2020learning}
Wei Wang, Junyu Gao, Xiaoshan Yang, and Changsheng Xu.
\newblock Learning coarse-to-fine graph neural networks for video-text retrieval.
\newblock \emph{IEEE Transactions on Multimedia}, 23:\penalty0 2386--2397, 2020.

\bibitem[Wang et~al.(2019)Wang, Wu, Chen, Li, Wang, and Wang]{wang2019vatex}
Xin Wang, Jiawei Wu, Junkun Chen, Lei Li, Yuan-Fang Wang, and William~Yang Wang.
\newblock Vatex: A large-scale, high-quality multilingual dataset for video-and-language research.
\newblock In \emph{Proceedings of the IEEE/CVF International Conference on Computer Vision}, pages 4581--4591, 2019.

\bibitem[Wang et~al.(2021)Wang, Zhu, and Yang]{wang2021t2vlad}
Xiaohan Wang, Linchao Zhu, and Yi Yang.
\newblock T2vlad: global-local sequence alignment for text-video retrieval.
\newblock In \emph{Proceedings of the IEEE/CVF Conference on Computer Vision and Pattern Recognition}, pages 5079--5088, 2021.

\bibitem[Wang et~al.(2022)Wang, Wu, Narasimhan, and Russakovsky]{wang2022multi}
Zeyu Wang, Yu Wu, Karthik Narasimhan, and Olga Russakovsky.
\newblock Multi-query video retrieval.
\newblock In \emph{Proceedings of European Conference on Computer Vision}, pages 233--249, 2022.

\bibitem[Wang et~al.(2023)Wang, Sung, Cheng, Bertasius, and Bansal]{wang2023unified}
Ziyang Wang, Yi-Lin Sung, Feng Cheng, Gedas Bertasius, and Mohit Bansal.
\newblock Unified coarse-to-fine alignment for video-text retrieval.
\newblock In \emph{Proceedings of the IEEE/CVF International Conference on Computer Vision}, pages 2816--2827, 2023.

\bibitem[Wray et~al.(2019)Wray, Larlus, Csurka, and Damen]{wray2019fine}
Michael Wray, Diane Larlus, Gabriela Csurka, and Dima Damen.
\newblock Fine-grained action retrieval through multiple parts-of-speech embeddings.
\newblock In \emph{Proceedings of the IEEE/CVF International Conference on Computer Vision}, pages 450--459, 2019.

\bibitem[Wray et~al.(2021)Wray, Doughty, and Damen]{wray2021semantic}
Michael Wray, Hazel Doughty, and Dima Damen.
\newblock On semantic similarity in video retrieval.
\newblock In \emph{Proceedings of the IEEE/CVF Conference on Computer Vision and Pattern Recognition}, pages 3650--3660, 2021.

\bibitem[Xiao et~al.(2025)Xiao, Hu, Li, and Hong]{xiao2025text}
Jian Xiao, Zhenzhen Hu, Jia Li, and Richang Hong.
\newblock Text proxy: Decomposing retrieval from a 1-to-n relationship into n 1-to-1 relationships for text-video retrieval.
\newblock In \emph{Proceedings of the AAAI Conference on Artificial Intelligence}, pages 8655--8663, 2025.

\bibitem[Xu et~al.(2016)Xu, Mei, Yao, and Rui]{xu2016msr}
Jun Xu, Tao Mei, Ting Yao, and Yong Rui.
\newblock Msr-vtt: A large video description dataset for bridging video and language.
\newblock In \emph{Proceedings of the IEEE/CVF Conference on Computer Vision and Pattern Recognition}, pages 5288--5296, 2016.

\bibitem[Xue et~al.(2023)Xue, Sun, Liu, Fu, Song, Li, and Luo]{xue2023clip}
Hongwei Xue, Yuchong Sun, Bei Liu, Jianlong Fu, Ruihua Song, Houqiang Li, and Jiebo Luo.
\newblock Clip-vip: Adapting pre-trained image-text model to video-language alignment.
\newblock In \emph{International Conference on Learning Representations}, 2023.

\bibitem[Yu et~al.(2025)Yu, Jiang, Dong, Gan, Yang, and Guo]{yu2025she}
Xuzheng Yu, Chen Jiang, Xingning Dong, Tian Gan, Ming Yang, and Qingpei Guo.
\newblock She-net: Syntax-hierarchy-enhanced text-video retrieval.
\newblock \emph{IEEE Transactions on Circuits and Systems for Video Technology}, 2025.

\bibitem[Yu et~al.(2018)Yu, Kim, and Kim]{yu2018joint}
Youngjae Yu, Jongseok Kim, and Gunhee Kim.
\newblock A joint sequence fusion model for video question answering and retrieval.
\newblock In \emph{Proceedings of European Conference on Computer Vision}, pages 471--487, 2018.

\bibitem[Zhai et~al.(2023)Zhai, Mustafa, Kolesnikov, and Beyer]{zhai2023sigmoid}
Xiaohua Zhai, Basil Mustafa, Alexander Kolesnikov, and Lucas Beyer.
\newblock Sigmoid loss for language image pre-training.
\newblock In \emph{Proceedings of the IEEE/CVF international conference on computer vision}, pages 11975--11986, 2023.

\bibitem[Zhang et~al.(2025)Zhang, Zeng, Gao, Song, Duan, Lyu, and Shen]{zhang2025text}
Haonan Zhang, Pengpeng Zeng, Lianli Gao, Jingkuan Song, Yihang Duan, Xinyu Lyu, and Heng~Tao Shen.
\newblock Text-video retrieval with global-local semantic consistent learning.
\newblock \emph{IEEE Transactions on Image Processing}, 34:\penalty0 3463--3474, 2025.

\bibitem[Zhao et~al.(2022)Zhao, Zhu, Wang, and Yang]{zhao2022centerclip}
Shuai Zhao, Linchao Zhu, Xiaohan Wang, and Yi Yang.
\newblock Centerclip: Token clustering for efficient text-video retrieval.
\newblock In \emph{Proceedings of the International ACM SIGIR Conference on Research and Development in Information Retrieval}, pages 970--981, 2022.

\bibitem[Zou et~al.(2022)Zou, Wu, Cheng, and Wang]{zou2022tokenflow}
Xiaohan Zou, Changqiao Wu, Lele Cheng, and Zhongyuan Wang.
\newblock Tokenflow: Rethinking fine-grained cross-modal alignment in vision-language retrieval.
\newblock \emph{arXiv preprint arXiv:2209.13822}, 2022.

\end{thebibliography}
}
\appendix

\clearpage

\setcounter{page}{1}
\maketitlesupplementary

\section{Fusion Module for Frames}\label{sec:app.video_pooling}
After obtaining text embedding $\vtext\in\mathbb{R}^{d}$ and frame embeddings $\vframes=[\rvf_1\|\rvf_1\|\cdots\|\rvf_M]^{\top}\in\mathbb{R}^{M\times d}$ for a text-video pair, we follow XPool~\ct{gorti2022x} to use $\vtext$ as condition, such that the fusion module $\pi(\vframes|\vtext)$ can attach more attention to the text-related frames.

Specifically, we employ cross-attention, where the text is the query and the frames are the key and value: 
\begin{small}
\begin{equation*}
    \begin{aligned}
        \mathbf{Q}_{t} &= \vtext^{\top}\mathbf{W}_{Q} \in \mathbb{R}^{1\times d_p}, \\
        \mathbf{K}_{F} &= \vframes\mathbf{W}_{K} \in \mathbb{R}^{M\times d_p}, \\
        \mathbf{V}_{F} &= \vframes\mathbf{W}_{V} \in \mathbb{R}^{M\times d_p}, 
    \end{aligned}
\end{equation*}
\end{small}
where $\mathbf{W}_{Q}$, $\mathbf{W}_{K}$, and $\mathbf{W}_{V}$ are learnable matrices, $d_p$ is the dimension of the cross-attention module. Then the global video presentation $\vvideo$ can be obtained as:
\begin{small}
\begin{equation*}
    \begin{aligned}
        \mathbf{z}&=\operatorname{LN}(\operatorname{Attention}(
            \mathbf{Q}_t, \mathbf{K}_F, \mathbf{V}_F)\mathbf{W}_O) \in \mathbb{R}^{1\times d}, \\
            \vvideo&=\operatorname{LN}(\operatorname{FC}(\mathbf{z})+
                \mathbf{z})^{\top} \in \mathbb{R}^{d}, 
    \end{aligned}
\end{equation*}
\end{small}
where $\operatorname{LN}(\cdot)$ is the layer-norm layer, and $\operatorname{FC}(\cdot)$ is a fully-connected layer. So we have $\pi(\vframes|\vtext) = \vvideo$.

\section{Stochastic Text Embeddings Sampling}\label{sec:app.sto_txt_modeling}

\subsection{Sampling}
Given text embedding $\vtext\in\mathbb{R}^{d}$ and frame embeddings $\vframes\in\mathbb{R}^{M\times d}$ for a text-video pair, we follow TMASS~\ct{wang2024text} to sample $S$ stochastic text embeddings $\{ \vtext_i^{\text{sto}} \}_{i=1}^{S}$ with extended semantics.

Here we only introduce how to sample one text embedding. Specifically, we first calculate a `radius' $\mathbf{r}$ of sampling:
\begin{small}
\begin{equation*}
    \begin{aligned}
        \mathbf{s} &= \operatorname{CosSim}(\vframes, \vtext) \in \mathbb{R}^{M}, \\
        \mathbf{r} &= \exp(\mathbf{s}^{\top}\mathbf{W}) \in \mathbb{R}^{d}, 
    \end{aligned}
\end{equation*}
\end{small}
where $\operatorname{CosSim}(\cdot,\cdot)$ is the consine similarity operation, and $\mathbf{W}\in \mathbb{R}^{M \times d}$ is a learnable matrix. Then the sampled text embedding is:
\begin{small}
\begin{equation*}
    \vtext_{i}^{\text{sto}} = \vtext + \mathbf{r}\odot\bm{\epsilon}_{i}, 
\end{equation*}
\end{small}
where $\bm{\epsilon}_{i}\sim\mathcal{N}(\mathbf{0}, \mathbf{1})$. Repeat for $S=20$ times, then we can obtain $S$ stochastic text embeddings $\{ \vtext_i^{\text{sto}} \}_{i=1}^{S}$. 

\subsection{Support Text}
Following~\ct{wang2024text}, we further obtain a support text $\rvt^{\text{sup}}$.
$\rvt^{\text{sup}}$ serves as the boundary of the randomness, within whose range the stochastic text embeddings are sampled. Including $\rvt^{\text{sup}}$ in the final loss can constrain the degree of random sampling.
\begin{small}
\begin{equation*}
    \vtext^{\text{sup}} = \vtext + \frac{\vvideo-\vtext}{\|\vvideo-\vtext\|}\|\mathbf{r}\|.
\end{equation*}
\end{small}

\section{Relational Graph Attention Network}\label{sec:app.rgat}

In \g (\cref{sec:frl}), for the text-frame graph with $n=1+S+M$ nodes, there are three types of edges (relations): text-text, text-frame, and frame-frame. 
So we have three adjacency matrices $\mathbf{A}^r \in \mathbb{R}^{n \times n}$ ($r \in \{ tt, ff, tf \}$), 
and $\mathbf{A}_{ij}^r=1$ when node pair $(i,j)$ is linked by relation $r$. 
For example, 
\begin{small}
\begin{equation}
    \mathbf{A}_{ij}^{tt} = \left\{
    \begin{aligned}
        &1, & i,j \in [1, 1+S],  \\
        &0, &\text{otherwise}.
    \end{aligned}
    \right.
\end{equation}
\end{small}

Once the attention scores $\alpha_{ij}^{r,h}$ is calculated (\cref{eq:attns_r}), 
we can construct node attention matrix $\mathbf{E}^{r,h}\in\mathbb{R}^{n \times n}$ as
\begin{small}
\begin{equation}
    \mathbf{E}_{ij}^{r,h} = \left\{
    \begin{aligned}
        &\alpha_{ij}^{r,h}, & \mathbf{A}_{ij}^{r} = 1, \\
        &0, & \mathbf{A}_{ij}^{r} = 0.
    \end{aligned}
    \right.
\end{equation}
\end{small}

So the second term in~\cref{eq:rgat} can be rewritten as
\begin{small}
\begin{equation}
\begin{aligned}
    &\Big\|_{h=1}^{H} \sum_r\sum_{j\in\mathcal{N}_i^{r}}\alpha_{ij}^{r,h}\mathbf{W}^{r,h}\mathbf{h}_j \\
    =&\Big\|_{h=1}^{H} \sum_r\sum_{j=1}^{n}\mathbf{E}_{ij}^{r,h}\mathbf{W}^{r,h}\mathbf{h}_j, 
\end{aligned}
\end{equation}
\end{small}
and it can be further written into matrix form as
\begin{small}
\begin{equation}
    \Bigg[ \Big\|_{h=1}^{H} \sum_r \mathbf{E}^{r,h}\mathbf{H}(\mathbf{W}^{r,h})^{\top} \Bigg]_i, \\
\end{equation}
\end{small}
where $\mathbf{H} = [\mathbf{h}_1\|\mathbf{h}_2\|\cdots\|\mathbf{h}_n]^{\top} \in \mathbb{R}^{n\times d}$ is the node matrix of the intermedia layer. 
Before the first layer, $\mathbf{H}$ is the initial node matrix $\mathbf{X}$. 
And ``$[\cdot]_{i}$" denotes the $i$-th row of the matrix.
Finally, the output node representation (\cref{eq:rgat}) can be rewritten into matrix form as
\begin{small}
\begin{equation}
    \mathbf{H}' = \sigma \Bigg( \mathbf{H}(\mathbf{W}_{\text{out}})^{\top} + 
    \Big\|_{h=1}^{H} \sum_r\mathbf{E}^{r,h}\mathbf{H}(\mathbf{W}^{r,h})^{\top} \Bigg). 
\end{equation}
\end{small}
Then $\mathbf{H}'$ is input of the next layer. 

In the final layer, the concatenation operation ``$\|_{h=1}^{H}$" is replaced by an average function across attention heads to obtain a $d$-dimentional output, and $\mathbf{W}_{\text{out}}\in\mathbb{R}^{d\times d}$.

\section{Training Energy-based Model}\label{sec:app.ebm}

\subsection{Derivations}
Given a distribution $p(x)$, energy-based model (\textbf{EBM}) models it by Boltzmann distribution $p_{\theta}(x)=\frac{\exp( -E_{\theta}(x) )}{Z_{\theta}}$. 
Its negative log-likelihood (NLL) loss is
\begin{small}
\begin{equation*}
    \loss_{\theta} = -\mathbb{E}_{x \sim p(x)}[\log p_{\theta}(x)].
\end{equation*}
\end{small}
To minimize $\loss_{\theta}$, we need to calculate the gradient of $\log p_{\theta}(x)$ w.r.t. $\theta$:
\begin{small}
\begin{equation*}
    \begin{aligned}
        &\nabla_{\theta} \log p_{\theta}(x) \\
        =& \nabla_{\theta} \log \frac{\exp( -E_{\theta}(x) )}{Z_{\theta}} \\
        =& -\nabla_{\theta} E_{\theta}(x) - \nabla_{\theta} \log Z_{\theta} \\
        =& -\nabla_{\theta} E_{\theta}(x) - \nabla_{\theta} \log \sum_x \exp( -E_{\theta}(x) ) \\
        =& -\nabla_{\theta} E_{\theta}(x) - \frac{1}{\sum_{x'} \exp( -E_{\theta}(x') )} \sum_x \nabla_{\theta} \exp( -E_{\theta}(x) ) \\
        =& -\nabla_{\theta} E_{\theta}(x) + \sum_x \frac{\exp( -E_{\theta}(x) )}{\sum_{x'} \exp( -E_{\theta}(x') )} \nabla_{\theta} E_{\theta}(x) \\
        =& -\nabla_{\theta} E_{\theta}(x) + \sum_x p_{\theta}(x) \nabla_{\theta} E_{\theta}(x) \\
        =& -\nabla_{\theta} E_{\theta}(x) + \mathbb{E}_{\Tilde{x}\sim p_{\theta}(\Tilde{x})}[ \nabla_{\theta} E_{\theta}(\Tilde{x}) ].
    \end{aligned}
\end{equation*}
\end{small}
So we have
\begin{small}
\begin{equation*}
    \begin{aligned}
        \nabla_{\theta} \loss_{\theta} &= -\nabla_{\theta} \mathbb{E}_{x \sim p(x)}[\log p_{\theta}(x)] \\
        &= \mathbb{E}_{x \sim p(x)}[ \nabla_{\theta} E_{\theta}(x) ] -  \mathbb{E}_{\Tilde{x} \sim p_{\theta}(\Tilde{x})}[ \nabla_{\theta} E_{\theta}(\Tilde{x}) ].
    \end{aligned}
\end{equation*}
\end{small}
Therefore, the EBM loss can be reformulated as
\begin{small}
\begin{equation*}
    \begin{aligned}
        \loss_{\theta} &= \mathbb{E}_{x \sim p(x)}[ E_{\theta}(x) ] -  \mathbb{E}_{\Tilde{x} \sim p_{\theta}(\Tilde{x})}[ E_{\theta}(\Tilde{x}) ] \\
        &\approx \frac{1}{B} \sum_{i=1}^B E_{\theta}(x_i) - \frac{1}{B'} \sum_{j=1}^{B'} E_{\theta}(\Tilde{x}_j), 
    \end{aligned}
\end{equation*}
\end{small}
where $\{\Tilde{x}_j\}_{j=1}^{B'}$ is $B'$ fake data sampled from $p_{\theta}(x)$. 
In practice, we set $B'=B$ for convenience.

\subsection{Sampling from EBM}
Given a Boltzmann distribution $p_{\theta}(x)=\frac{\exp( -E_{\theta}(x) )}{Z_{\theta}}$, the gradient of $\log p_{\theta}(x)$ w.r.t. $x$ is
\begin{small}
\begin{equation*}
    \begin{aligned}
        \nabla_x \log p_{\theta}(x) &= \nabla_x \log \frac{\exp( -E_{\theta}(x) )}{Z_{\theta}} \\
        &= -\nabla_x E_{\theta}(x) - \nabla_x \log Z_{\theta} \\
        &= -\nabla_x E_{\theta}(x).
    \end{aligned}
\end{equation*}
\end{small}
So we can first initialize a sample using uniform distribution $x^{(0)} \sim \mathcal{U}$, 
and utilize multi-step stochastic gradient ascending to make the sample follows the distribution $p_{\theta}(x)$:
\begin{small}
\begin{equation*}
    \begin{aligned}
        x^{(k+1)} &= x^{(k)} + \eta \nabla_{x^{(k)}} \log p_{\theta} \left(x^{(k)} \right) + \epsilon^{(k)} \\
        &= x^{(k)} - \lambda \nabla_{x^{(k)}} E_{\theta} \left(x^{(k)} \right) + \epsilon^{(k)},
    \end{aligned}
\end{equation*}
\end{small}
where $\epsilon \sim \mathcal{N}\left(0, \sigma^2 \right)$. After $K$ steps, $x^{(K)}$ is the output.

\subsection{Replay Buffer}
Furthermore, we use sample \textbf{relay buffer}~\ct{du2019implicit}. 
The initialization of MCMC chain plays a crucial role in mixing time, but Langevin dynamics does not place restrictions on sample initialization given sufficient sampling steps. 
Thus, we use a sample replay buffer $\mathcal{R}$ in which we preserve previously generated samples and use either these samples or uniform noise for initialization.

\subsection{Regularization for EBM Loss}
Additionally, we adopt the $L_2$ regularization for energy magnitudes of both true data and sampled data when computing $\loss_{\text{ebm}}$ during training, 
as otherwise while the difference between true data and sampled data was preserved, the actual values would fluctuate to numerically unstable values~\ct{du2019implicit}.
For an energy function $E_{\theta}(\cdot)$, true data $x\sim p(x)$, and sampled data $\Tilde{x}\sim p_{\theta}(\Tilde{x})$:
\begin{small}
\begin{equation*}
    \begin{aligned}
        \loss_{\text{ebm}} = &\frac{1}{B} \sum_{i=1}^{B} E_{\theta}(x_i) - \frac{1}{B'} \sum_{j=1}^{B'} E_{\theta}(\Tilde{x}_j) \\
        &+c \Bigg[ \frac{1}{B} \sum_{i=1}^{B} E_{\theta}(x_i)^2 + \frac{1}{B'} \sum_{j=1}^{B'} E_{\theta}(\Tilde{x}_j)^2 \Bigg],
    \end{aligned}
\end{equation*}
\end{small}
where $c$ is a coefficient. We set $c=1.0$.

\subsection{Algorithm for Sampling Text-Frame Pairs}

To calculate the second term in energy loss (\cref{eq:eam_loss}), a $K$-step MCMC sampling is conducted to sample $B'$ text-frame pairs using energy function $E_{\theta}$. 
At the beginning of training, the replay buffers for text and frames are initialized as empty set:
\begin{small}
\begin{equation*}
    \mathcal{R}_t \leftarrow \varnothing, \ \ \ 
    \mathcal{R}_F \leftarrow \varnothing, 
\end{equation*}
\end{small}
and the whole sampling algorithm is shown in~\cref{algo:sampling_from_ebm}.

\begin{algorithm}[!t]
\caption{Sampling Algorithm of \e.}
\label{algo:sampling_from_ebm}

\textbf{Input}: Energy function $E_{\theta}(\cdot,\cdot)$, replay buffers $\mathcal{R}_t$ and $\mathcal{R}_F$.\\
\textbf{Parameter}: MCMC step size $\eta$, MCMC noise variance $\sigma^2$, MCMC steps $K$, number of MCMC samples $B'$.\\
\textbf{Output}: Sampled text-frame embedding pairs $\{\Tilde{\vtext}_j,\Tilde{\vframes}_j\}_{i=1}^{B'}$, and the updated replay buffers $\mathcal{R}_t$ and $\mathcal{R}_F$.

\begin{algorithmic}[1]

\FOR{$j = 1$ {\bfseries to} $B'$}
    \STATE \small{$\Tilde{\vtext}_j^{(0)} \sim \mathcal{R}_t$} with $95\%$ probability and $\mathcal{U}$ otherwise; 
    \STATE \small{$\Tilde{\vframes}_j^{(0)} \sim \mathcal{R}_F$} with $95\%$ probability and $\mathcal{U}$ otherwise; 
    \FOR{$\text{sample step } k = 1$ {\bfseries to} $K$}
        \STATE \small{$\eps_{t_j}^{(k)} \sim \mathcal{N}(0, \sigma^2)$}; 
        \STATE \small{$\Tilde{\vtext}_j^{(k)} = \Tilde{\vtext}_j^{(k-1)} - \eta \nabla_{\Tilde{\vtext}_j^{(k-1)}}E_{\theta}(\Tilde{\vtext}_j^{(k-1)},\Tilde{\vframes}_j^{(k-1)}) + \eps_{t_j}^{(k)}$};
        \STATE \small{$\eps_{v_j}^{(k)} \sim \mathcal{N}(0, \sigma^2)$}; 
        \STATE \small{$\Tilde{\vframes}_j^{(k)} = \Tilde{\vframes}_j^{(k-1)} - \eta \nabla_{\Tilde{\vframes}_j^{(k-1)}}E_{\theta}(\Tilde{\vtext}_j^{(k-1)},\Tilde{\vframes}_j^{(k-1)}) + \eps_{v_j}^{(k)}$};
    \ENDFOR
    \STATE \small{$\Tilde{\vtext}_j \leftarrow \texttt{stop-gradient}\big[\Tilde{\vtext}_j^{(K)}\big]$}; 
    \STATE \small{$\mathcal{R}_t \leftarrow \mathcal{R}_t \cup \{\Tilde{\vtext}_j\}$}; 
    \STATE \small{$\Tilde{\vframes}_j \leftarrow \texttt{stop-gradient}\big[\Tilde{\vframes}_j^{(K)}\big]$}; 
    \STATE \small{$\mathcal{R}_F \leftarrow \mathcal{R}_F \cup \{\Tilde{\vframes}_j\}$}; 
\ENDFOR
\STATE Sampled text-frame embedding pairs $\{\Tilde{\vtext}_j,\Tilde{\vframes}_j\}_{i=1}^{B'}$; 
\STATE Updated replay buffers $\mathcal{R}_t$ and $\mathcal{R}_F$. 

\end{algorithmic}
\end{algorithm}

\section{Experiment Settings}\label{sec:app.exp_settings}

\subsection{Datasets}
We evaluate the proposed \shortname using four widely used datasets.

(1) \textbf{MSRVTT}~\ct{xu2016msr} includes 10,000 video clips, each of which has 20 captions. We employ the 1k-A split~\ct{liu2019use,gabeur2020multi} to use 9k videos for training and 1k videos for validation and test. 

(2) \textbf{DiDeMo}~\ct{anne2017localizing} contains 10,642 Flickr video clips and 40,543 captions. We follow~\ct{lin2022text} to concatenate all captions of a video as a query. 

(3) \textbf{MSVD}~\ct{chen2011collecting} is composed of 1,970 video clips with a length ranging from 1 to 62 seconds, each of which contains about 40 captions. We split the training, validation, and test sets with 1,200, 100, and 670 videos following~\ct{luo2022clip4clip}. 

(4) \textbf{VATEX}~\ct{wang2019vatex} consists of 34,991 video clips, each of which is related to multiple captions. We apply the same dataset split as~\ct{chen2020fine} to select 25,911, 1,500, and 1,500 videos as training, validation, and test sets. 

\subsection{Evaluation Metrics}
Following~\ct{luo2022clip4clip}, we use the standard TVR metrics, including Recall (\%) at rank 1/5/10 (R@1, R@5, R@10), Median Rank (MdR), Mean Rank (MnR), and Rsum (R@1$+$R@5$+$R@10). We also evaluate the average training time per epoch (sec./batch) and the total test time (sec.) on MSRVTT with ViT-B/16. 

\subsection{Selected baselines}
We select three types of models as baselines for comparison. 
The first type is early works that are based on pre-extracted text and video features (not end-to-end models), including CE~\ct{liu2019use}, MMT~\ct{gabeur2020multi}, HiT~\ct{liu2021hit}, and TeachText~\ct{croitoru2021teachtext}. 
The second group is the end-to-end models that use pre-trained networks (\eg, ResNet~\ct{he2016deep} and CLIP) as encoders, including ClipBert~\ct{lei2021less}, SupportSet~\ct{patricksupport}, Fronzen~\ct{bain2021frozen}, CLIP4Clip~\ct{luo2022clip4clip}, TS2Net~\ct{liu2022ts2}, CLIP-VIP~\ct{xue2023clip}, XPool~\ct{gorti2022x}, TMASS~\ct{wang2024text}, RAP~\ct{cao2024rap}, TFVL~\ct{shen2025temporal}, GLSCL~\ct{zhang2025text}, TempMe~\ct{shentempme}, TV-ProxyNet~\ct{xiao2025text}, and Video-ColBERT~\ct{reddy2025video}. 
And our proposed \shortname is an end-to-end model using pre-trained CLIP encoders as the backbone. 

\subsection{Implementation Details}
The pre-trained CLIP model (ViT-B/32 and ViT-B/16) is used as image and text encoder~\cite{radford2021learning}. 
We use CLIP4Clip~\cite{luo2022clip4clip} as backbone, and apply similar configurations, such that the hidden dimension $d=512$, weight decay $0.2$, dropout $0.3$, learning rate $10^{-7}$ for CLIP modules and $10^{-4}$ for non-CLIP modules; and we use a cosine scheduler for warmup, with the proportion of $0.1$. 
We uniformly sample $M=12$ frames for each video clip on all datasets ($64$ for DiDeMo). All frames are resized into $224 \times 224$. We set the max length of text to $32$ for all datasets ($64$ for DiDeMo). 
For \g, the number of text candidates $S=20$, the number of attention heads in RGAT $H=4$, and the number of RGAT layers $L=2$. 
For MCMC sampling in \e, step number $K=20$, step size $\eta=1$, and noise variance $\sigma^2=0.005$. 
The sigmoid loss is initialized as $\tau_p=4.77$ and $b=-12.93$ following~\cite{zhai2023sigmoid}. 
For loss coefficient, we set $\lambda_{\text{sup}}=0.8$ following~\cite{wang2024text}, and $\lambda_{\text{eam}}=1.0$ by default. 
We train all models on two A100 GPUs for $5$ epochs with a total batch size of $B=64$. 
In inference, we run on one A100 GPU with a batch size of $8$. 

\section{Additional Analysis}
\label{sec:app.additional_analysis}

\begin{table}[!t]
\centering

\resizebox{0.65\linewidth}{!}{

\begin{tabular}{lr|r|r}
\multicolumn{2}{c|}{\textbf{Model}} & \multicolumn{1}{c|}{\textbf{Training}} & \multicolumn{1}{c}{\textbf{Test}} \\
\hline
\multicolumn{2}{l|}{CLIP4Clip} & 2947.4 & 49.7 \\
\multicolumn{2}{l|}{XPool} & 3062.7 & 51.0 \\
\multicolumn{2}{l|}{TS2Net} & 4367.1 & 99.7 \\
\multicolumn{2}{l|}{CLIP-VIP} & 2497.0 & 41.1 \\
\multirow{2}{*}{UCoFiA} & (t2v) & \multirow{2}{*}{21021.4} & 2969.8 \\
 & (v2t) &  & 10000.3 \\
\multicolumn{2}{l|}{ProST} & 3199.1 & 51.4 \\
\multicolumn{2}{l|}{TMASS} & 3066.1 & 115.3 \\
\multicolumn{2}{l|}{GLSCL} & 4078.2 & 148.5 \\
\multicolumn{2}{l|}{TempMe} & 5271.7 & 18.2 \\
\multicolumn{2}{l|}{TV-ProxyNet} & 2619.1 & 74.2 \\
\multicolumn{2}{l|}{Video-ColBERT} & 3542.3 & 44.0 \\
\multicolumn{2}{l|}{\shortname (Ours)} & 5207.0 & 141.7
\end{tabular}

}

\vspace{-5pt}
\caption{Average training time per epoch and test time (sec.) consumption of different models (ViT-B/16) on MSRVTT.}
\label{tab:time}

\end{table}

\subsection{Complexity Analysis}
The main differences between our proposed \shortname and other baselines lie in three components in the text-video similarity computation stage: (1) stochastic text embeddings sampling; (2) Fine-Grained Relationship Learning (\g); (3) Energy-Aware Matching (\e). 

\paragraph{\textbf{Stochastic Text Embeddings Sampling.}} Following TMASS~\ct{wang2024text}, we first calculate text-frame similarity and use a linear layer to determine the radius, so the complexity is given by $\mathcal{O}(B^2M^2d+B^2Md)=\mathcal{O}(B^2M^2d)$, where $B$ is batch size, $M$ is number of frames, and $d$ is the number of latent dimensions. 
Since we need to consider each text-video pair during the similarity computation, so the number of pairs is $B^2$. Then we use this radius to sample $S$ text embeddings, which adds a complexity of $\mathcal{O}(B^2Sd)$. The total complexity remains $\mathcal{O}(B^2M^2d)$ since $M^2 >> S$.

\paragraph{\textbf{Fine-Grained Relationship Learning.}} This process primarily uses a $L$-layer $H$-head RGAT to process text-frames graph, which contain $n=1+S+M$ nodes. In each layer, RGAT first transforms the inputs, calculates the attention matrix, and obtain the output node features(\cref{eq:eij_r,eq:attns_r,eq:rgat}), yielding a complexity of $\mathcal{O}(B^2RHnd^2+B^2RHnd+B^2RHn^2d)=\mathcal{O}(B^2RHnd^2)$, where $R$ is the number of relationship types. 
The final text embedding (\cref{eq:text_frame_weights,eq:text_nodes_attns,eq:text_gen}) is computed with a complexity of $\mathcal{O}(B^2Sd)$. 
So the complexity of each RGAT layer is $\mathcal{O}(B^2RHnd^2)$, such that the final complexity is $\mathcal{O}(B^2LRHnd^2)$. 

\paragraph{\textbf{Energy-Aware Matching.}} In this part, the $K$-step MCMC sampling incurs the most time cost, which we focus on primarily. In each sample step, we use a 2-layer MLP as the energy function to calculate the scalar energy of a text-video pair and to calculate the gradient (\cref{eq:efunc_mlp,eq:mcmc_sampling}), so the total complexity is $\mathcal{O}(K{B'}Md{d'}+K{B'}M{d'})=\mathcal{O}(K{B'}Md{d'})$, where ${B'}$ is the number of sampled text-video pairs (normally we set ${B'}=B$), and ${d'}$ is the number of latent dimensions of energy function. 
Since we consider the energy of each text-video pair, the number of samples is $B$. 
So the total complexity of \e is $\mathcal{O}(KBMd{d'})$. 

\paragraph{}In the training phase, the overall additional complexity compared to other baselines is the summation of all three components above, which is $\mathcal{O}(B^2LRHnd^2)$. Since \e is only conducted during training, the complexity for the test phase is the combination of the first and second components, which also results in $\mathcal{O}(B^2LRHnd^2)$. Overall, the additional computational overhead of \shortname primarily comes from \g and \e. This remains manageable in practice because $B$, $L$, $R$, $H$, and $n$ are typically small, and $d$ is a fixed constant. This indicates that our model incurs only modest additional runtime compared to baselines (\cref{tab:time}).

\subsection{Simple Alternative for Graph Encoder}

As illustrated in~\cref{sec:frl}, the constructed text-frame graph is a heterogeneous graph with multiple types of nodes and edges, 
we proposed to use RGAT (\cref{eq:rgat}) for learning the edge attention scores. 
If we ignore the types of links and regard the text-frame graph as a fully connected graph, a simple GAT can be used as an alternative.

Specifically, we simply treat the text-frame graph as a fully-connected graph, \ie, $\mathbf{A}_{ij} \equiv 1$, so the edge weight is given by
\begin{small}
\begin{align}
    e_{ij}^{h} &= \psi \bigg( \left[\mathbf{W}^{h}\mathbf{h}_i \| \mathbf{W}^{h}\mathbf{h}_j \right] \bigg) \in \mathbb{R}, 
\end{align}
\end{small}
then the attention score is calculated as
\begin{small}
\begin{align}
    \alpha_{ij}^{h} &= \frac{ \exp \left(\operatorname{LeakyReLU}\left(e_{ij}^{h} \right)\right) }{ \sum_{k=1}^{n}\exp\left(\operatorname{LeakyReLU}\left(e_{ik}^{h} \right)\right) }, 
\label{eq:attns}
\end{align}
\end{small}
where $n=1+S+M$ is the number of nodes, and we can construct the node attention matrix $\mathbf{E}^h=\{ \alpha_{ij}^{h} \}_{ij}\in\mathbb{R}^{n \times n}$.

The output node matrix of each layer is
\begin{small}
\begin{equation}
    \mathbf{H}' = \sigma \left( \Big\|_{h=1}^{H}\mathbf{E}^{h}\mathbf{H}(\mathbf{W}^{h})^{\top} \right). 
\label{eq:gat}
\end{equation}
\end{small}
In the final layer, the concatenation operation ``$\|_{h=1}^{H}$" is replaced by an average function following Appendix~\ref{sec:app.rgat}.

Similarly, we select the edge weights from \textbf{the final layer} and average them across attention heads to get the final text-frame weights, \ie, 
\begin{small}
\begin{equation}
    e_{ij} = \frac{1}{H} \sum_{h=1}^{H}e_{ij}^{h}, \ \  
    \begin{aligned}
        &i \in [1, 1+S], \\
        &j \in [1+S+1, 1+S+M], 
    \end{aligned}
\end{equation}
\end{small}

\begin{table}[!t]
\centering

\resizebox{\linewidth}{!}{

\begin{tabular}{c|ccc|ccc}
\textbf{Graph} & \multicolumn{3}{c|}{\textbf{MSRVTT}} & \multicolumn{3}{c}{\textbf{DiDeMo}} \\
\textbf{Encoder} & \textbf{R@1↑} & \textbf{R@5↑} & \textbf{R@10↑} & \textbf{R@1↑} & \textbf{R@5↑} & \textbf{R@10↑} \\
\hline
\multicolumn{7}{c}{w/o \e} \\
\hline
GAT & 49.4 & \underline{76.8} & \textbf{85.9} & 47.5 & 76.6 & 85.9 \\
RGAT & \underline{49.9} & \textbf{77.3} & 84.3 & \underline{49.1} & 77.5 & 85.0 \\
\hline
\multicolumn{7}{c}{w/ \e} \\
\hline
GAT & \underline{49.9} & 76.6 & 85.2 & 48.9 & \underline{78.4} & \underline{86.6} \\
\rowcolor{gray!15}
RGAT & \textbf{51.0} & 76.2 & \underline{85.6} & \textbf{51.5} & \textbf{78.7} & \textbf{86.8}
\end{tabular}

}

\vspace{-5pt}
\caption{Text-to-video on different graph encoders.}
\label{tab:ablaion-genc}

\end{table}

From~\cref{tab:ablaion-genc}, Overall, RGAT performs better than GAT on both MSRVTT and DiDeMo with or without \e. 
This illustrates that RGAT is better at learning multi-relationships than GAT. 
Moreover, the performance will improve when employing \e, further demonstrating the effectiveness of our proposed \e in enhancing \g and the final performance. 
We select RGAT as the graph encoder in the final implementation.

\subsection{Ablation Study for Coefficient of Loss Function}
For the loss coefficient, 
the setting of $\lambda_{sup}=0.8$ follows the practice in TMASS~\cite{wang2024text}, 
and the coefficient $\lambda_{eam}$ was set to 1 by default to maintain a stable and consistent gradient contribution from EAM. 
We further conduct a sensitivity analysis on the loss coefficients. 
Since $\lambda_{sup}=0.8$ has been validated as an effective choice for balancing supervision signals for balancing supervision signals~\cite{wang2024text}, we focus our analysis on $\lambda_{eam}$.  

From~\cref{tab:ablaion-loss_coef}, the performance remains stable across various $\lambda_{eam}$, and the best overall performance is achieved when $\lambda_{eam}=1.0$. 
This indicates that the proposed model is not sensitive to the choice of $\lambda_{eam}$, which supports our default setting for simplicity and robustness.

\begin{table}[!t]
\centering

\resizebox{\linewidth}{!}{

\begin{tabular}{c|ccc|ccc}
\multirow{2}{*}{$\lambda_{eam}$} & \multicolumn{3}{c|}{\textbf{MSRVTT}} & \multicolumn{3}{c}{\textbf{DiDeMo}} \\
 & \textbf{R@1↑} & \textbf{R@5↑} & \textbf{R@10↑} & \textbf{R@1↑} & \textbf{R@5↑} & \textbf{R@10↑} \\
\hline
0.1 & \underline{50.7} & \textbf{76.6} & \underline{85.7} & 50.9 & \textbf{78.9} & \underline{86.6} \\
0.5 & 50.1 & \underline{76.5} & \textbf{86.2} & \underline{51.2} & 78.6 & \underline{86.6} \\
\rowcolor{gray!15}
1.0 & \textbf{51.0} & 76.2 & 85.6 & \textbf{51.5} & \underline{78.7} & \textbf{86.8}
\end{tabular}

}

\vspace{-5pt}
\caption{Text-to-video on different $\lambda_{eam}$.}
\label{tab:ablaion-loss_coef}

\end{table}

\subsection{Ablation Study for Frame-to-Frame Edges}

To verify the effectiveness of modeling internal frame-to-frame relationships within a video, we evaluate a variant of \shortname that removes only the frame-to-frame edges (f2f) in \g. 

As shown in~\cref{tab:ablaion-wo_f2f}, comparing the ``w/o f2f" variant with our full \shortname reveals a consistent performance improvement across both datasets (e.g., +0.4\% R@1 on MSRVTT and +1.6\% R@1 on DiDeMo). 
This gain directly validates the importance of modeling inter-frame dependencies, as it enables the RGAT to capture temporal continuity and the logical progression of video content, rather than treating frames as independent, isolated entities.

\begin{table}[!t]
\centering

\resizebox{\linewidth}{!}{

\begin{tabular}{l|ccc|ccc}
 & \multicolumn{3}{c|}{\textbf{MSRVTT}} & \multicolumn{3}{c}{\textbf{DiDeMo}} \\
\hline
 & \textbf{R@1↑} & \textbf{R@5↑} & \textbf{R@10↑} & \textbf{R@1↑} & \textbf{R@5↑} & \textbf{R@10↑} \\
w/o \g & 48.6 & 75.3 & 83.7 & 43.8 & 75.6 & 83.8 \\
w/o f2f & \underline{50.6} & \textbf{77.5} & \underline{84.6} & \underline{49.9} & \textbf{78.7} & \underline{86.5} \\
\rowcolor{gray!15}
\shortname & \textbf{51.0} & \underline{76.2} & \textbf{85.6} & \textbf{51.5} & \textbf{78.7} & \textbf{86.8}
\end{tabular}

}

\vspace{-5pt}
\caption{Text-to-video on frame-to-frame edges (f2f).}
\label{tab:ablaion-wo_f2f}

\end{table}

\subsection{Visualization}

\subsubsection{Different Models}\label{sec:app.visualization-models}
\begin{figure*}
    \centering
    \includegraphics[width=\linewidth]{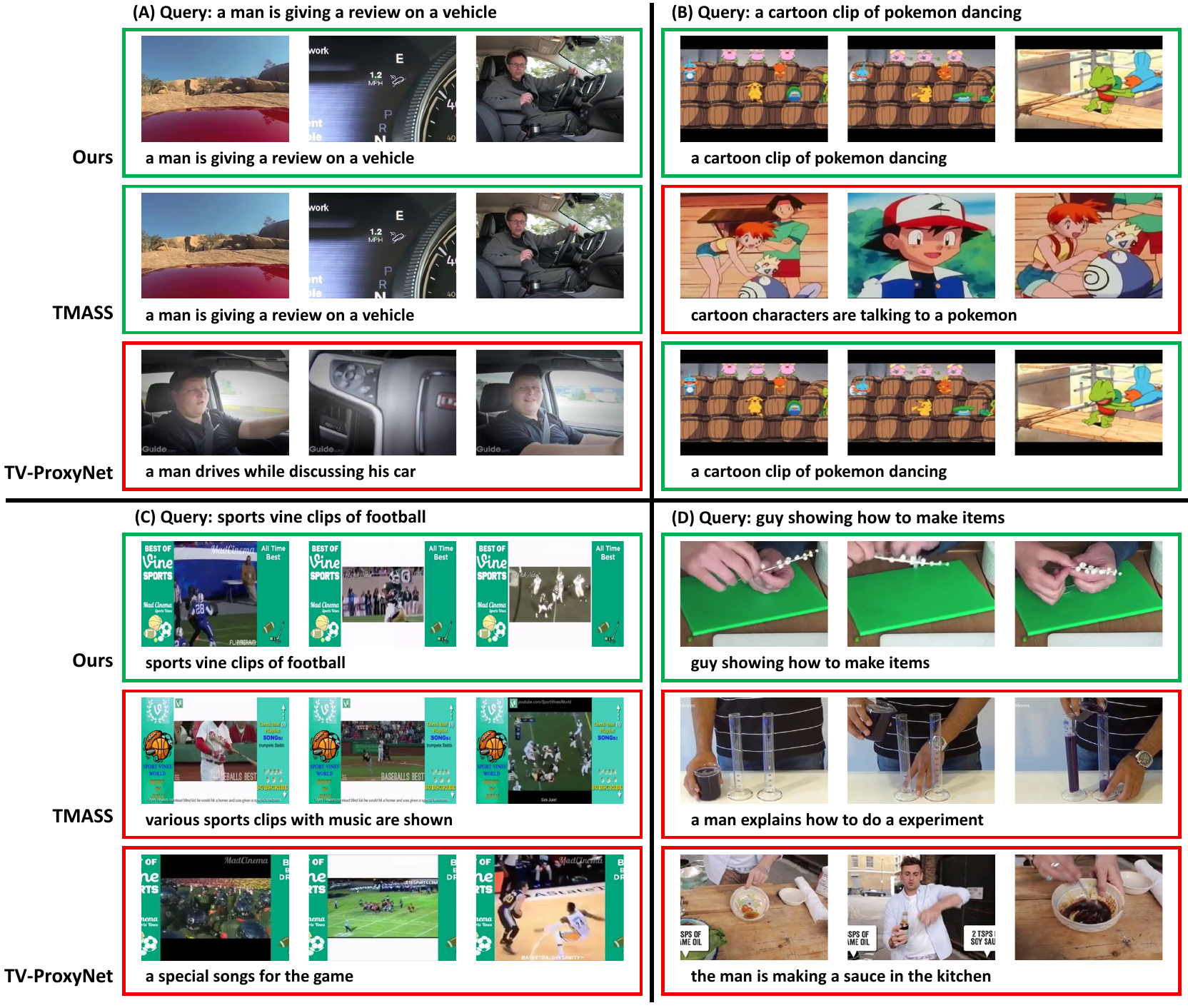}
    \vspace{-15pt}
    \caption{Visualization of text-to-video retrieval results by our proposed \shortname, TMASS~\ct{wang2024text}, and TV-ProxyNet~\ct{xiao2025text}. 
    \textcolor{green!70!black}{Green} denotes the correct retrieval result, and \textcolor{red}{red} is the wrong result.}
    \label{fig:visualization}
\end{figure*}

In \cref{fig:visualization} we visualize some text-to-video retrieval results by our proposed \shortname, TMASS~\ct{wang2024text}, and TV-ProxyNet~\ct{xiao2025text}. 

TMASS and TV-ProxyNet do not exploit the internal frame-to-frame relationships and the video context, so they might retrieve incorrect videos with similar semantics but different contexts. For example, in \cref{fig:visualization} (A), TV-ProxyNet retrieves the video where there is a man and a car, ignoring that the man is driving the car but not giving a review. 
Similarly, in~\cref{fig:visualization} (B), TMASS retrieves the video with Pokémon, but they are not dancing. 

Furthermore, as previously mentioned, the enriched text embeddings generated by TMASS can negatively impact retrieval outcomes due to the inherent randomness from stochastic sampling. However, \shortname mitigates this individual bias.
As illustrated in ~\cref{fig:visualization}, \shortname successfully retrieves the correct video clip, while TMASS retrieves an incorrect clip with similar semantics. For instance, given the query text ``sports vine clips of football," \shortname retrieves the appropriate video, whereas TMASS provides a video showcasing various sports, including baseball and American football.
This highlights the accuracy of our generated text embeddings, whereas TMASS introduces some undesirable information into the embeddings.

\subsubsection{Different Components}\label{sec:app.visualization-components}
\begin{figure*}
    \centering
    \includegraphics[width=\linewidth]{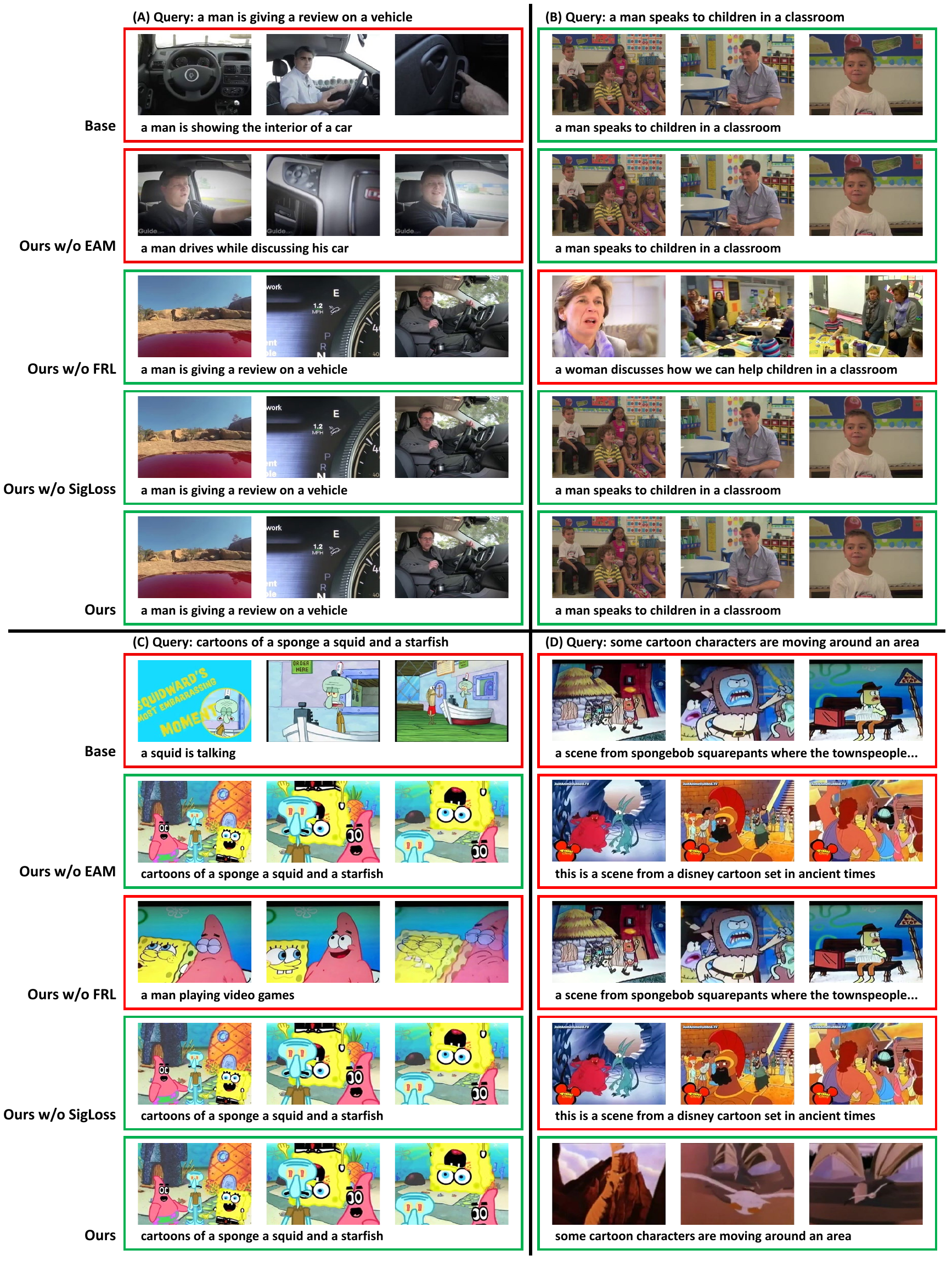}
    \vspace{-15pt}
    \caption{Visualization of text-to-video retrieval results by different components of \shortname. 
    \textcolor{green!70!black}{Green} denotes the correct retrieval result, and \textcolor{red}{red} is the wrong result.}
    \label{fig:visualization-ablation}
\end{figure*}

In \cref{fig:visualization-ablation} we visualize results by different components, baseline, without \g, without \e, and our proposed \shortname. In the absence of one component, the model might lose some key information, leading to a similar but incorrect video. 

For example, the enriched text cannot capture frame contextual information when not without \g (`Base' and `Ours w/o \g'). 
In \cref{fig:visualization-ablation} (B), the “w/o \g” variant only captures the elements “speaks” and “children in a classroom” in the returned video, while ignoring the mismatch between the “woman” in the video and the “man” mentioned in the query.
Similarly, in \cref{fig:visualization-ablation} (C), both “Base” and “w/o FRL” retrieve videos as long as one or two of the mentioned characters appear, without considering whether all the characters mentioned in the query are actually present throughout the video.

\end{document}